\documentclass[10pt,twocolumn,letterpaper]{article}

\usepackage{iccv}
\usepackage{times}
\usepackage{epsfig}
\usepackage{graphicx}
\usepackage{amsmath}
\usepackage{amssymb}

\usepackage[numbers,sort&compress]{natbib}
\usepackage{subcaption}
\usepackage{booktabs}
\usepackage{makecell}
\usepackage{supertabular}

\usepackage[accsupp]{axessibility}

\usepackage{hyperref}
\hypersetup{pagebackref=true,breaklinks=true,letterpaper=true,colorlinks,bookmarks=false}

\iccvfinalcopy

\ificcvfinal\fi

\usepackage[usenames,dvipsnames]{xcolor}
\usepackage{url}
\usepackage{multirow}
\usepackage{colortbl}
\usepackage{pifont}
\usepackage{paralist}
\usepackage{nicematrix}

\newcommand{\cmark}{\ding{51}}%
\newcommand{\xmark}{\ding{55}}%

\newcommand{\figref}[1]{Fig.~\ref{#1}}
\newcommand{\eqnref}[1]{Eq.~(\ref{#1})}
\newcommand{\tabref}[1]{Tab.~\ref{#1}}
\newcommand{\secref}[1]{Sec.~\ref{#1}}
\newcommand{\supplementary}{Supp. Mat.\xspace}

\newcommand{\vectr}[1]{\boldsymbol{#1}}

\DeclareMathOperator*{\argmin}{arg\,min}

\newcommand{\datasetname}{EMDB\xspace}
\newcommand{\datasetnamefull}{Electromagnetic Database of Global 3D Human Pose and Shape in the Wild\xspace}
\newcommand{\datasetnamefullhighlighted}{\textbf{E}lectro\textbf{M}agnetic \textbf{D}ata\textbf{B}ase of Global 3D Human Pose and Shape in the Wild\xspace}
\newcommand{\methodname}{EMP\xspace}
\newcommand{\methodnamefull}{Electromagnetic Poser\xspace}

\begin{document}

\title{
EMDB: The Electromagnetic Database of Global 3D Human \\Pose and Shape in the Wild
}

\author{Manuel Kaufmann\textsuperscript{1} \quad Jie Song\textsuperscript{1} \quad Chen Guo\textsuperscript{1} \quad Kaiyue Shen\textsuperscript{1} \quad Tianjian Jiang\textsuperscript{1} \\ Chengcheng Tang\textsuperscript{2} \quad Juan José Zárate\textsuperscript{1} \quad Otmar Hilliges\textsuperscript{1} \\
    \\
    \textsuperscript{1}ETH Zürich, Department of Computer Science \quad 
    \textsuperscript{2}Meta Reality Labs\\
}

\ificcvfinal\thispagestyle{empty}\fi

\twocolumn[{
            \renewcommand\twocolumn[1][]{#1}%
            \maketitle
            \vspace{-3em}
            \begin{center}
        	\centering
        	\includegraphics[width=1.0\textwidth]{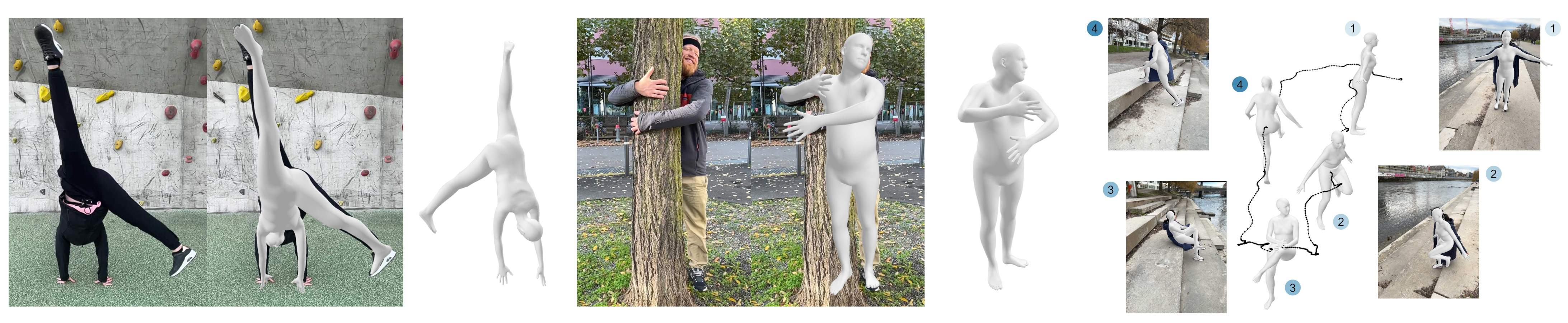}
        	\captionof{figure}{\datasetname is a novel dataset that provides accurate SMPL pose and shape parameters including global camera and body trajectories for in-the-wild videos. (left and middle) Challenging example poses taken from \datasetname. (right, 1-4) Visualization of global body center trajectory and the corresponding 3D poses projected into the camera view.}
        	\label{fig:teaser}
            \end{center}
        }]

\ificcvfinal\thispagestyle{empty}\fi

\begin{abstract}
    \vspace{-0.5cm}
   We present \datasetname, the \datasetnamefull. 
   \datasetname is a novel dataset that contains high-quality 3D SMPL pose and shape parameters with global body and camera trajectories for in-the-wild videos.
   We use body-worn, wireless electromagnetic (EM) sensors and a hand-held iPhone to record a total of 58 minutes of motion data, distributed over 81 indoor and outdoor sequences and 10 participants. Together with accurate body poses and shapes, we also provide global camera poses and body root trajectories.
   To construct \datasetname, we propose a multi-stage optimization procedure, which first fits SMPL to the 6-DoF EM measurements and then refines the poses via image observations.
   To achieve high-quality results, we leverage a neural implicit avatar model to reconstruct detailed human surface geometry and appearance, which allows for improved alignment and smoothness via a dense pixel-level objective.
   Our evaluations, conducted with a multi-view volumetric capture system, indicate that \datasetname has an expected accuracy of 2.3 cm positional and 10.6 degrees angular error, surpassing the accuracy of previous in-the-wild datasets.
   We evaluate existing state-of-the-art monocular RGB methods for camera-relative and global pose estimation on \datasetname.
   \datasetname is publicly available under \href{https://ait.ethz.ch/emdb}{https://ait.ethz.ch/emdb}.
\end{abstract}

\section{Introduction}

3D human pose and shape estimation from monocular RGB images is a long-standing computer vision problem with many applications in AR/VR, robotics, assisted living, rehabilitation, or sports analysis.
Much progress has been made in estimating camera-relative poses, typically assuming a weak-perspective camera model, \eg, \cite{Li2021hybrik, Sun2021ROMP, Kocabas2021PARE,li2022cliff,Cho2022FastMETRO}. However, this setting is too restrictive for many applications that involve a moving camera.  
Such applications must estimate a) human poses in-the-wild, under occlusion and encountering uncommon poses; and b) global locations of humans and the camera.
Compared to the camera-relative setting, there is relatively little work on global pose estimation \cite{yuan2022glamr,Ye2023SLAHMR}. 
This is in part due to the lack of comprehensive datasets that contain accurate 3D human pose and shape with global trajectories in a fully in-the-wild setting. 

To overcome this bottleneck, in this paper we propose a novel dataset, called \datasetname, short for the \datasetnamefullhighlighted.
\datasetname consists of 58 minutes (105k frames) of challenging 3D human motion recorded in diverse scenes. We provide high-quality pose and shape annotations, as well as global body root and camera trajectories. The dataset contains 81 sequences distributed over 10 participants that were recorded with a hand-held mobile phone. 

Recording such data requires a motion capture system that is both mobile and accurate -- a notoriously difficult problem.
Systems that provide world-anchored 3D body keypoints often require multiple well-calibrated RGB or IR cameras within a static environment, which restricts outdoor use \cite{joo2015panoptic,Mehta2017MPI,h36m_pami,Huang2022RICH}.
While body-worn sensors such as head-mounted cameras \cite{rhodin2016egocap,Zhang2022EgoBody,xu2019mo2cap2} are promising for mobile use, such egocentric approaches introduce either heavy self-occlusions \cite{rhodin2016egocap,xu2019mo2cap2} or are restricted to indoor settings with a fixed capture volume \cite{Zhang2022EgoBody}.
The 3DPW dataset \cite{vonMarcard20183dpw} uses IMU sensors for outdoor recordings, yet the dataset is relatively small and lacks global trajectories. 
Moreover, IMU drift and the lack of direct positional sensor measurements imposes constraints in terms of pose diversity and accuracy.
Instead, following \cite{kaufmann2021empose}, we leverage drift-free electromagnetic (EM) sensors that directly measure their position and orientation.
Yet, any sensor-based capture system requires handling of measurement noise, accurate calibration of the sensors to the body's coordinate system and temporal and spatial alignment of the data streams.

Addressing these challenges, we propose a method, \methodnamefull (\methodname), that allows for the construction of \datasetname.
\methodname is a multi-stage optimization formulation that fuses up to 12 body-worn EM sensor measurements, monocular RGB-D images and camera poses, and produces accurate SMPL \cite{SMPL:2015} pose and shape parameters alongside global trajectory estimates for the body's root and the camera. \methodname works in the following 3 stages.

\noindent{\textbf{Calibration and EM Pose:}}
As an initial calibration step, we scan participants in minimal clothing using an indoor multi-view volumetric capture system (MVS, \cite{MRCS}) to obtain ground-truth shape and skin-to-sensor offsets.
We subsequently record in-the-wild sequences of the same subject and fit SMPL to the drift-free EM measurements of the sensors' positions and orientations.
This provides an accurate SMPL fit, albeit in a EM-local coordinate system.

\noindent{\textbf{World Alignment:}}
In the second stage, we align the EM-local pose estimates with a global world space, defined by the tracking space of a hand-held iPhone 13 that films the participants.
We model this stage as a joint optimization that fuses the input EM measurements, 2D keypoints, depth, and camera poses.
In our experiments we have found that the self-localized 6D poses of the iPhone are accurate to around $2$ cm positional and $< 1$ degree angular error.
The fixed body shape and accurate camera poses thus enable \methodname to provide global SMPL root trajectories.

\noindent{\textbf{Pixel-Level Refinement:}}
In the third stage, we refine the initial global poses via dense pixel-level information to ensure high-quality and temporally smooth image alignment.
To this end we leverage recent advancements in neural body modelling for in-the-wild videos and fit a neural body model with detailed geometry and appearance to the RGB images.
Following~\cite{Guo2023v2a}, we model the human as a deformable implicit signed distance field and the background as a neural radiance field.
This allows us to formulate a pixel-level RGB loss that compares color values obtained via composited neural rendering with the observed pixel value.
We jointly optimize the neural body model and the SMPL poses, initialized with the output of the second stage.
We experimentally show that this final stage results in temporally smooth results and accurate pose-to-image alignment.

We evaluate \methodname on 21 sequences recorded with our MVS \cite{MRCS}, the same system we use to register ground-truth SMPL shape parameters. With a pose accuracy of 2.3 cm positional and 10.6\textdegree~angular error, our evaluations reveal that \methodname is more accurate than what has been reported for 3DPW (2.6 cm, 12.1\textdegree)~\cite{vonMarcard20183dpw}.
Also,  our global SMPL root trajectories are accurate with an estimated error of 5.1 cm compared to our indoor MVS.
Finally, we evaluate the performance of recent state-of-the-art camera-relative and global RGB-based pose estimators on \datasetname.
Our results show that \datasetname is a new challenging dataset that will enable future local \emph{and} global pose estimation research.

In summary, we contribute:
\begin{inparaenum}
\item \datasetname, to the best of our knowledge the first comprehensive dataset to provide accurate SMPL poses, shapes, and trajectories in an unrestricted, mobile, in-the-wild scenario.
    \item \methodname, the first method to fuse EM measurements with image data and camera poses.
    \item Extensive evaluations of the accuracy of \methodname as well as baseline results of state-of-the-art work when evaluating on \datasetname.
\end{inparaenum}
Data is available under \href{https://ait.ethz.ch/EMDB}{https://ait.ethz.ch/emdb}.

\section{Related Work}

\paragraph*{Sensor-based Pose Estimation}
Modern inertial measurement units (IMUs) are an appealing sensor modality for human pose estimation because they are small and do not require line-of-sight. However, they only measure orientation directly. This lack of reliable positional information can be mitigated by using a large number of sensors \cite{roetenberg2007moven} or by fusing IMU data with other modalities such as external cameras \cite{Mallesonreal, trumble2017total, ponsmollCVPR2010, pons2011outdoor, Gilbert2019Fusing, zhe2020fusingimu, vonMarcard20183dpw,Bleser2011Ego}, head-mounted cameras \cite{Guzov2021HPS}, LiDAR \cite{Dai2022HSC4D}, or acoustic sensors \cite{Vlasic2007,liu2011realtime}.
Research has attempted to reduce the required number of sensors, \eg \cite{von2017sparse,huang2018dip, vonMarcard20183dpw,Butt2021IMU}, which requires costly optimizations \cite{von2017sparse}, external cameras \cite{vonMarcard20183dpw}, or data-driven priors to establish the sensor-to-pose mapping \cite{huang2018dip, Jiang2022TransformerInertialPoser, Yi2022PIP,Yi2021TransPose,Winkler2022QuestSim} and deal with the under-constrained pose space.
While such methods yield accurate local poses, IMUs are intrinsically limited in that their position estimates drift over time.

Addressing this challenge, EM-POSE \cite{kaufmann2021empose} puts forth a novel method for body-worn pose estimation that relies on wireless electromagnetic (EM) field sensing to directly measure positional values. A learned optimization \cite{song2020human} formulation estimates accurate body pose and shape from EM inputs.
However, \cite{kaufmann2021empose} is limited to a small indoor capture space, requires external tracking of the root pose and is not aligned with image observations. In this work, we move beyond these limitations and present an EM-based capture system that is mobile, deployed to capture in-the-wild data, and produces high-quality pose-to-image alignment.

\paragraph*{RGB-based Pose Estimation}
The 3D pose of a human is either represented as a skeleton of 3D joints \cite{martinez2017simple,mehta2017vnect,zhou2016sparseness,sun2018integral}  or via parametric body models like SCAPE~\cite{anguelov2005scape} and SMPL~\cite{SMPL:2015} for a more fine-grained representation.
We note that almost the entire body of research estimates local (\ie, camera-local) poses. 
In recent years, deep neural networks have driven significant advancements in estimating body model parameters directly from images or videos \cite{kanazawa2018end,tung2017self,tan2018indirect,omran2018neural,guler2019holopose,varol2017learning,xu2019denserac,zheng2019deephuman, Li2021hybrik, kocabas2019vibe,Kocabas2021PARE,pymaf2021,Sun2021ROMP,li2022cliff,Sun2022BEV,joo2020eft}. In addition, researchers have combined the advantages of both optimization and regression to fit the SMPL body \cite{kolotouros2019learning, song2020human}.
Others have leveraged graph convolutional neural networks to effectively learn local vertex relations by building a graph structure based on the mesh topology of the parametric body models, \eg \cite{Cho2022FastMETRO,lin2021-mesh-graphormer}. These methods propose transformer encoder architectures to learn the non-local relations between human body joints and mesh vertices via attention mechanisms. 
Recently, a few approaches have set out to estimate realistic global trajectories of humans and cameras from local human poses \cite{yuan2022glamr,li2022dnd,yu2021human,Ye2023SLAHMR}. We evaluate several of the above methods on our proposed dataset on the tasks of camera-relative and global human pose estimation.

\paragraph*{Human Pose Datasets}

Commonly used datasets to evaluate 3D human pose estimation are H3.6M \cite{h36m_pami}, MPI-INF-3DHP \cite{Mehta2017MPI}, HumanEva \cite{Sigal2010HumanEva}, and TotalCapture \cite{joo2015panoptic}. Although these datasets offer synchronized video and MoCap data, they are restricted to indoor settings with static backgrounds and limited variation in clothing and activities.

To address these limitations, \cite{vonMarcard20183dpw} proposed a method that combines a single hand-held camera and a set of body-worn IMUs to estimate relatively accurate 3D poses, resulting in an in-the-wild dataset called 3DPW. Following this work, HPS \cite{Guzov2021HPS} estimates 3D human pose with IMUs while localizing the person via a head-mounted camera within a pre-scanned 3D scene. To further address the issue of IMU drift, HSC4D \cite{Dai2022HSC4D} leverages LiDAR sensors for global localization. However, both HPS and HSC4D assume static scene scans and do not register global body pose in a third-person view.
Moreover, they lack an evaluation of how accurate their pose estimates are.
Another approach to outdoor performance capture with reduced equipment is to utilize one or multiple RGB-D cameras \cite{bhatnagar22behave, Huang2022RICH, Hassan2019PROX}.
In these approaches, the quality of body pose registrations is limited by the cameras' line-of-sight, noisy depth measurements and the capture space is fixed. 
None of these works provide an estimate of their datasets' accuracy either.
EgoBody \cite{Zhang2022EgoBody} provides egocentric views and registered SMPL poses but is restricted to a fixed indoor space, requires up to 5 external RGB-D cameras and lacks evaluation of the data accuracy.
Synthetic data has been suggested as a means to provide high-quality annotations \cite{Patel2021AGORA,varol2017learning}. However, due to the reliance on static human scans and artificial backgrounds there is a distributional shift compared to real images.

With \datasetname we provide the first dataset of 3D human pose and shape that is recorded in an unrestricted, mobile, in-the-wild setting and provides global camera and SMPL root trajectories. To gauge the accuracy of \datasetname, we rigorously evaluate our method against ground-truth obtained on a multi-view volumetric capture system \cite{MRCS}. These evaluations reveal that \datasetname is not only two times larger than 3DPW, but its annotations are also more accurate.

\section{Overview}

\begin{figure*}[t]
    \includegraphics[width=\linewidth, trim={0 0 0 0},clip]{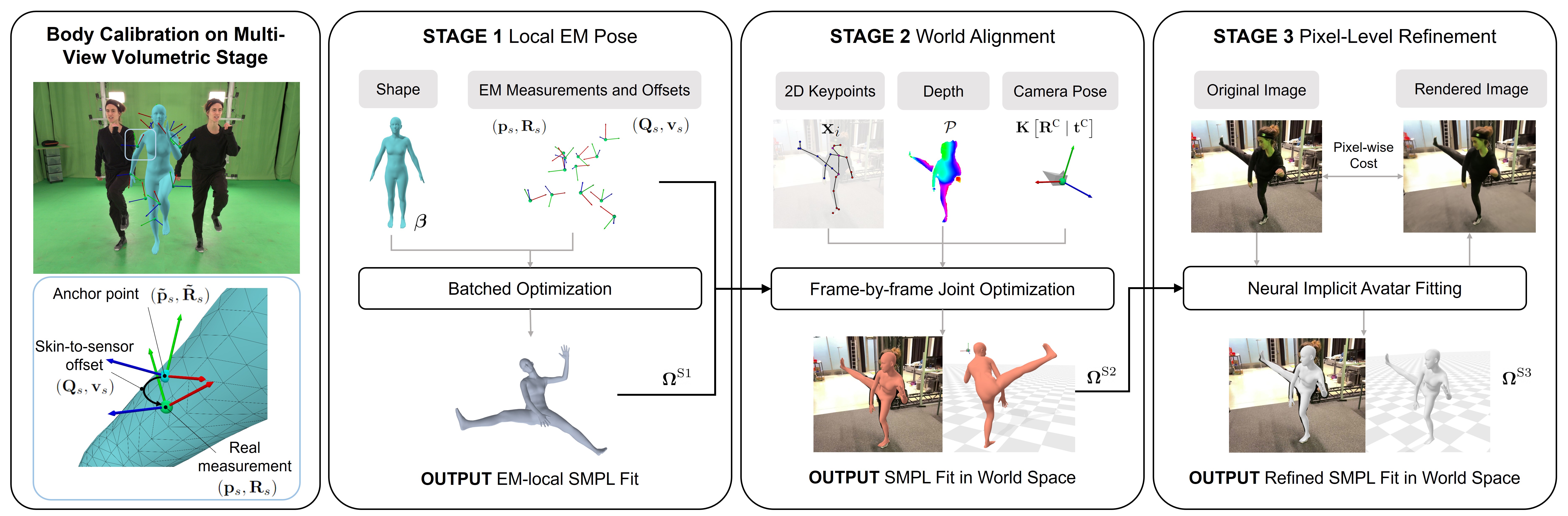}
    \caption{
    Method overview. We first scan a subject in minimal clothing with a multi-view volumetric capture system to obtain their reference shape parameters $\vectr{\beta}$ and calibrate subject-specific skin-to-sensor offsets in regular clothing (left). We subsequently fit SMPL to in-the-wild data with a multi-stage optimization pipeline. Stage 1 fits SMPL to the EM measurements in EM-local space leveraging the calibrated body shape and skin-to-sensor offsets. Stage 2 aligns the local fit with the world, by jointly optimizing over 2D keypoints, depth, camera poses, EM measurements, and the output of stage 1. Stage 3 then refines the output of stage 2 by fitting a neural implicit body model with detailed geometry and appearance to the RGB images via a pixel-level supervision signal to boost smoothness and image-to-pose alignment.}
    \label{fig:overview}
\end{figure*}

Our goal is to provide a dataset with i) accurate 3D body poses and ii) shapes alongside global trajectories of the iii) body's root and iv) the moving camera. This data is obtained from electromagnetic (EM) sensor measurements and RGB-D data streamed from a single hand-held iPhone.
We first describe the capture setup and protocol in \secref{sec:capture_setup}.
\secref{sec:method} discusses our method, \methodname, for the estimation of global SMPL parameters, summarized in \figref{fig:overview}.
To gauge the accuracy of \methodname, we evaluate it against ground-truth data recorded with a multi-view volumetric system (MVS, \cite{MRCS}). These evaluations are provided in \secref{sec:evaluation}.
Finally, using \methodname on newly captured in-the-wild sequences, we introduce the \datasetnamefull, \datasetname, in \secref{sec:emdb}, where we also evaluate existing state-of-the-art methods on \datasetname.

\section{Capture Setup}
\label{sec:capture_setup}
\subsection{Sensing Hardware}
EM sensors measure their position $\mathbf{p}_s$ and orientation $\mathbf{R}_s$ w.r.t.~a source that emits an electromagnetic field.
We use the same wireless EM sensors as \cite{kaufmann2021empose}, which have an estimated accuracy of 1 cm positional and 2-3 degrees angular error.
We mount the EM source on the lower back of a participant and arrange the sensors on the lower and upper extremities and the head and torso. For the detailed sensor placement we refer to the \supplementary
All sensor data is streamed wirelessly to a laptop for recording.

We record the subjects with a hand-held iPhone 13 Pro Max. The record3d app \cite{record3d} is used to retrieve depth and the iPhone's 6D pose is estimated by Apple's ARKit.
We synchronize the data streams via a hand clap which is easy to detect in the phone's audio and in the EM  accelerations.

\subsection{Body Calibration}
\label{sec:body_calibration}
Before we start recording, we first scan each participant in minimal clothing to obtain their ground-truth shape.
To this end, we leverage our MVS \cite{MRCS} and use the resulting surface scans and 53 RGB views to register the SMPL shape parameters $\vectr{\beta}$.
Details on the registration pipeline can be found in the \supplementary

Subsequently, we mount the sensors and EM-source onto the participant under regular clothing (see \figref{fig:overview}, left). We then record a 3-second calibration sequence to determine subject-specific skin-to-sensor offsets.
We first register SMPL to the calibration sequence and follow \cite{kaufmann2021empose} to manually select anchor points on the SMPL mesh for every sensor $s$. An anchor point is parameterized via a position $\mathbf{\tilde{p}}_s$ and orientation $\mathbf{\tilde{R}}_s$. We then compute per-sensor offsets $\mathbf{o}_s = (\mathbf{Q}_s, \mathbf{v}_s)$ by minimizing an objective that equates the measured orientation $\mathbf{R}_s = \mathbf{\tilde{R}}_s \mathbf{Q}_s$ and the measured position $\mathbf{p}_s = \mathbf{\tilde{p}}_s + \mathbf{\tilde{R}}_s\mathbf{v}_s$ (see \figref{fig:overview}, left).
For this to work, the sensor measurements must be spatially and temporally aligned with the MVS. We thus track the EM source with an Apriltag \cite{olson2011apriltag1, wang2016apriltag2, krogius2019apriltag3} and use an Atomos Ultrasync One timecode generator \cite{atomosTimecode} for temporal alignment.
More details are shown in the \supplementary Note that this procedure must only be done once per sensor placement. 

\section{Method (\methodname)}
\label{sec:method}

\subsection{Notations and Preliminaries}
The inputs to our method are EM sensor measurements $\mathbf{p}_s \in \mathbb{R}^3$ and $\mathbf{R}_s \in SO(3)$, skin-to-sensor offsets $\mathbf{o}_s = (\mathbf{Q}_s, \mathbf{v}_s)$, SMPL shape parameters $\vectr{\beta} \in \mathbb{R}^{10}$, RGB images $\mathbf{I} \in \mathbb{R}^{1920 \times 1440 \times 3}$, depth point clouds $\mathcal{P} = \{ \mathbf{p}_i \mid \mathbf{p}_i \in \mathbb{R}^3 \}$, camera extrinsics $\mathbf{C} = \left[ \mathbf{R}^\text{C} \mid \mathbf{t}^\text{C} \right] \in \mathbb{R}^{3 \times 4}$ and intrinsics $\mathbf{K} \in \mathbb{R}^{3 \times 3}$.
Note that the EM measurements are in EM-local space, \ie, relative to the source worn on the lower back.
From these input measurements, we aim to estimate the SMPL body pose parameters $\vectr{\theta}_b \in \mathbb{R}^{69}$, the SMPL root orientation $\vectr{\theta}_r \in \mathbb{R}^{3}$ and translation $\mathbf{t} \in \mathbb{R}^{3}$ in world coordinates such that they align with sensor measurements, images, and camera poses.
We fix the world space to be the iPhone's coordinate frame. We summarize SMPL parameters as $\vectr{\Omega} = (\vectr{\theta}_r, \vectr{\theta}_b, \mathbf{t}, \vectr{\beta}$). 
Note that $\vectr{\beta} \in \mathbb{R}^{10}$ is not an optimization variable and is obtained a-priori (see \secref{sec:body_calibration}). 
All quantities usually refer to a time step $t$, but we omit the time subscript for clarity unless necessary.

\subsection{Multi-stage Optimization}

As shown in \figref{fig:overview}, our method employs a multi-stage optimization procedure, which we detail in the following.

\paragraph*{Stage 1: Local EM Pose}
For a given sequence, we start our optimization procedure by first finding SMPL parameters $\vectr{\Omega}$ that best explain the EM measurements in EM-local space.
We follow EM-POSE \cite{kaufmann2021empose} and define a reconstruction cost function $E_{\text{rec}}$ that measures how well the current SMPL fit matches the sensor measurements:
\begin{align}
    E_{\text{rec}} =
        &\sum_{s=1}^S \lambda_\text{p} || \mathbf{p}_s - \mathbf{p}_s^v(\mathcal{M}(\vectr{\Omega}), \mathbf{o}_s)||^2_2 + \nonumber \\
        &\sum_{s=1}^S \lambda_\text{r}|| \mathbf{R}_s - \mathbf{R}_s^v(\mathcal{M}(\vectr{\Omega}), \mathbf{o}_s)||^2_2  \ ,
    \label{eq:e_rec}
\end{align}
\noindent where we use the current SMPL mesh $\mathcal{M}(\vectr{\Omega})$ and skin-to-sensor offsets $\mathbf{o}_s$ to compute virtual sensor positions $\mathbf{p}_s^v$ and orientations $\mathbf{R}_s^v$.
In addition, we penalize impossible joint angles with a simple regularizer $E_\text{bp}$.
The final optimization objective of the first stage is then $E_\text{S1} = \lambda_\text{rec} E_\text{rec} + \lambda_\text{bp}E_\text{bp}$.
We use a batched optimization to minimize it over all $T$ frames of the sequence. The output of stage 1 are the SMPL parameters in local EM space, $\mathbf{\Omega}^\text{S1}$ (see also \figref{fig:overview}).

\paragraph*{Stage 2: World Alignment}
Due to accurate sensor data and our body calibration procedure, the $\mathbf{\Omega}^\text{S1}$ parameters are already of high quality (see \secref{sec:evaluation_gt}).
However, the EM space is not aligned with the world space. We align $\mathbf{\Omega}^\text{S1}$ with the world in a second optimization stage such that it fits the RGB-D observations and camera pose data. An overview of this stage is provided in \figref{fig:overview}.

This stage is guided by a 2D keypoint reprojection loss. 
Importantly, both 2D keypoints and depth are noisy and fitting to them na\"ively can corrupt the initial estimates $\vectr{\Omega}^\text{S1}$.
Hence, we must trade-off accurate alignment of human and camera poses in world coordinates with the accuracy of the local pose.
Although our trust in the EM fit $\vectr{\Omega}^\text{S1}$ is high, we can still achieve improvements by fitting to RGB-D data for frames in which errors arise from sensor calibration or occasional measurement noise. 
Furthermore, the temporal alignment of EM and RGB-D data streams can be improved by fitting to the images.
We model this trade-off as a joint optimization over all the input modalities.

We first define a 2D keypoint reprojection loss. We extract $N=25$ 2D keypoints from Openpose \cite{cao2017OpenposeRealtime} denoted by $\mathbf{x}_i \in \mathbb{R}^{2}$. The 3D keypoints $X(\vectr{\Omega})$ are obtained via a linear regressor from the SMPL vertices.
We then use the camera parameters to perspectively project the 3D keypoints (in homogenous coordinates), $\hat{\mathbf{x}}_i = \mathbf{K} \left[ \mathbf{R}^\text{C} \mid \mathbf{t}^\text{C}  \right] X(\vectr{\Omega})_i$. 
The reprojection cost is then defined as
\begin{equation}
    E_\text{2D} = \sum_{i=1}^N \mathbb{I} \left[c_i \geqslant \tau\right] \cdot \rho( \hat{\mathbf{x}}_i - \mathbf{x}_i)
    \label{eq:e_2d}
\end{equation}
\noindent where $\rho$ is the Geman-McClure function \cite{Geman1987StatisticalMF}, $c_i$ is the confidence of the $i$-th keypoint as estimated by Openpose and $\mathbb{I}$ the indicator function.
We set a high confidence threshold $\tau = 0.5$ in \eqnref{eq:e_2d} to account for keypoint noise. Yet, even high confidence keypoints can be wrong. To ensure high quality of the ground-truth annotations provided in \datasetname, we carefully review the keypoint predictions by Openpose and manually correct them for challenging samples.

We add two EM-related cost terms to this stage's optimization to further constrain the 3D pose.
The first term is the EM reconstruction cost $E_\text{rec}$ from \eqnref{eq:e_rec}. Note that here we only optimize the SMPL body pose $\vectr{\theta}_b$ when computing the cost, denoted as $E^*_\text{rec}$.
The second term is an additional prior on the body pose $\vectr{\theta}^\text{S1}_b$ found in the first stage:
\begin{equation}
    E_\text{prior} = || \vectr{\theta}^\text{S1}_b - \vectr{\theta}_b\||^2_2 .
\end{equation}
This $E_\text{prior}$ formulation is similar to the one of HPS~\cite{Guzov2021HPS}. However, we found that the addition of $E_\text{prior}$ alone is not sufficient and $E^*_\text{rec}$ plays a crucial role (see \secref{sec:evaluation_gt}).

Finally, we incorporate the iPhone's point clouds $\mathcal{P}$. Since the point clouds are noisy, they mostly serve as a regularizer for the translation $\mathbf{t}$ with the following term:
\begin{equation}
    E_\text{pcl} = \frac{1}{|\mathcal{P}^h|} \sum_{\mathbf{p}_i \in \mathcal{P}_h} d(\mathbf{p}_i, \mathcal{M}(\vectr{\Omega})) .
    \label{eq:e_pcl}
\end{equation}
Here, $d(\cdot)$ finds the closest triangle on the SMPL mesh $\mathcal{M}(\vectr{\Omega})$ and then returns the squared distance to either the triangle's plane, edge, or vertex. $\mathcal{P}_h$ is a crop of $\mathcal{P}$, where the human is isolated via masks provided by RVM \cite{Lin2021RVM}.
The final second stage objective is thus:
\begin{align}
    E_\text{S2} = \lambda_\text{2D}E_\text{2D} + \lambda_\text{rec}E^*_\text{rec} + \lambda_\text{prior}E_\text{prior} + \lambda_\text{pcl}E_\text{pcl}
\end{align}

We optimize this objective frame-by-frame and use the previous output as the initialization for the next frame. The output of this stage is $\vectr{\Omega}^\text{S2}$ (see also \figref{fig:overview}). For the very first frame, we initialize $\mathbf{t}^\text{S2}$ as the mean of $\mathcal{P}_h$. All sequences start with a T-pose where the subject is facing the camera, so that it is easy to find an initial estimate of $\vectr{\theta}_r^\text{S2}$.

\paragraph*{Stage 3: Pixel-Level Refinement}
Stage 2 finds a good trade-off between accurate poses and global alignment (see \secref{sec:evaluation_gt}).
However, the jitter in the 2D keypoints causes temporally non-smooth estimates.
Reducing the jitter by manually cleaning 2D keypoints is not viable.
Instead, we add a third stage to \methodname (see also \figref{fig:overview}) in which we follow recent developments in neural body modelling for in-the-wild videos. For every sequence, we fit a neural implicit model of clothed human shape and appearance to the RGB images by minimizing a dense pixel-level objective.

More specifically, we leverage Vid2Avatar (V2A \cite{Guo2023v2a}) to model the human in the scene as an implicit signed-distance field (SDF) representing surface geometry and a texture field, while the background is treated as a separate neural radiance field (NeRF++) \cite{https://doi.org/10.48550/arxiv.2010.07492}.
The SDF is modelled in canonical space and deformed via SMPL parameters $\vectr{\Omega}$ to pose the human.
Then, given a ray $\mathbf{r} = (\mathbf{o}, \mathbf{v})$ whose origin $\mathbf{o}$ is the camera center and $\mathbf{v}$ its viewing direction, a color value $C(\mathbf{r})$ can be computed via differentiable neural rendering and is compared to the actual RGB value $\hat{C}(\mathbf{r})$ to formulate a self-supervised objective:

\begin{equation}
    E_\text{rgb} = \frac{1}{|\mathcal{R}_t|}\sum_{\mathbf{r} \in \mathcal{R}_t} |C(\mathbf{r}) - \hat{C}(\mathbf{r}) |
\end{equation}

\noindent where $\mathcal{R}_t$ is the set of all rays that we shoot into the scene at frame $t$.
Importantly, $C(\mathbf{r})$ depends on the SMPL poses $\vectr{\Omega}$ that are optimized jointly together with the parameters for the human and background fields.
Along with $E_\text{rgb}$, the original formulation of V2A minimizes two other objectives: the Eikonal loss $E_\text{eik}$ and a scene decomposition loss $E_\text{dec}$ to disentangle the human from the background.
For more details we refer the reader to \cite{Guo2023v2a}.
We initialize the SMPL parameters $\vectr{\Omega}$ with the outputs of the second stage $\vectr{\Omega}^\text{S2}$ and add a pose regularization term $E_\text{reg} = || \vectr{\theta} - \vectr{\theta}^\text{S2} ||^2_2$ (where $\vectr{\theta} := \left[ \vectr{\theta}_r, \vectr{\theta}_b \right])$ to encourage solutions to stay close to the initializations.
The final third stage objective for a single time step is thus (omitting weights $\lambda$ for brevity):
\begin{equation}
    E_\text{S3} = E_\text{rgb}(\vectr{\omega}_h, \vectr{\omega}_b) + E_\text{eik}(\vectr{\omega}_h) + E_\text{dec}(\vectr{\omega}_h) + E_\text{reg}(\vectr{\theta}) ,
\end{equation}
\noindent where $\vectr{\omega}_h$ summarizes the parameters for the human field, including SMPL pose parameters $\vectr{\Omega}$, and $\vectr{\omega}_b$ summarizes the weights of the background field. This objective is minimized over all $T$ frames of the given sequence and produces outputs $\vectr{\Omega}^\text{S3}$, which are noticeably less jittery (see \secref{sec:evaluation_gt}).
\section{Evaluation}
\label{sec:evaluation}

\begin{table}
\centering
\resizebox{1.00\linewidth}{!}{
\begin{tabular}{lccc}
\toprule
Method & MPJPE-PA & MPJAE-PA & Jitter \\
& [mm] & [deg] & [10m s$^{-3}$] \\ \hline
ROMP \cite{Sun2021ROMP} & 57.9 $\pm$ 23.6 & 19.8 $\pm$ 6.3 & 49.0 $\pm$ 10.6\\
HybrIK \cite{Li2021hybrik} & 50.4 $\pm$ 22.3 & 19.0 $\pm$ 5.8 & 33.3 $\pm$ 7.1 \\
Vid2Avatar \cite{Guo2023v2a} & 50.2 $\pm$ 22.8 & 18.1 $\pm$ 6.2 & 38.7 $\pm$ 8.0 \\
LGD \cite{song2020human} & 61.1 $\pm$ 31.9 & 20.1 $\pm$ 8.0 & 68.9 $\pm$ 10.2 \\
\hline
Stage 1 & 26.0 $\pm$ 8.6 & 10.9 $\pm$ 3.1 & 6.0 $\pm$ 2.9 \\
Stage 2 (no $E^*_\text{rec}$) & 31.6 $\pm$ 14.1 & 12.7 $\pm$ 4.5 & 26.8 $\pm$ 3.7\\
Stage 2 (no $E_\text{prior}$) & 35.4 $\pm$ 14.2 & 11.6 $\pm$ 3.9 & 23.0 $\pm$ 3.3\\
Stage 2 & 23.7 $\pm$ \textbf{7.5} & \textbf{10.5} $\pm$ \textbf{3.0} & 21.7 $\pm$ 3.7 \\ \hline
Stage 3 (after $E_\text{S3}$) & 23.5 $\pm$ 7.6 & 10.6 $\pm$ 3.1 & 12.7 $\pm$ 2.5\\
Stage 3 (\methodname) & \textbf{23.4} $\pm$ \textbf{7.5} & 10.6 $\pm$ 3.1 & \textbf{3.5} $\pm$ \textbf{1.0}\\
\bottomrule
\end{tabular}
}
\caption{
Comparison of \methodname to existing RGB-based methods (top) and self-ablations (middle/bottom) on ground-truth data obtained with our multi-view capture system.}
\label{tab:evaluation_gt}
\end{table}

\begin{figure}[t]
    \includegraphics[width=\linewidth, trim={0 0 0 1.3cm},clip]{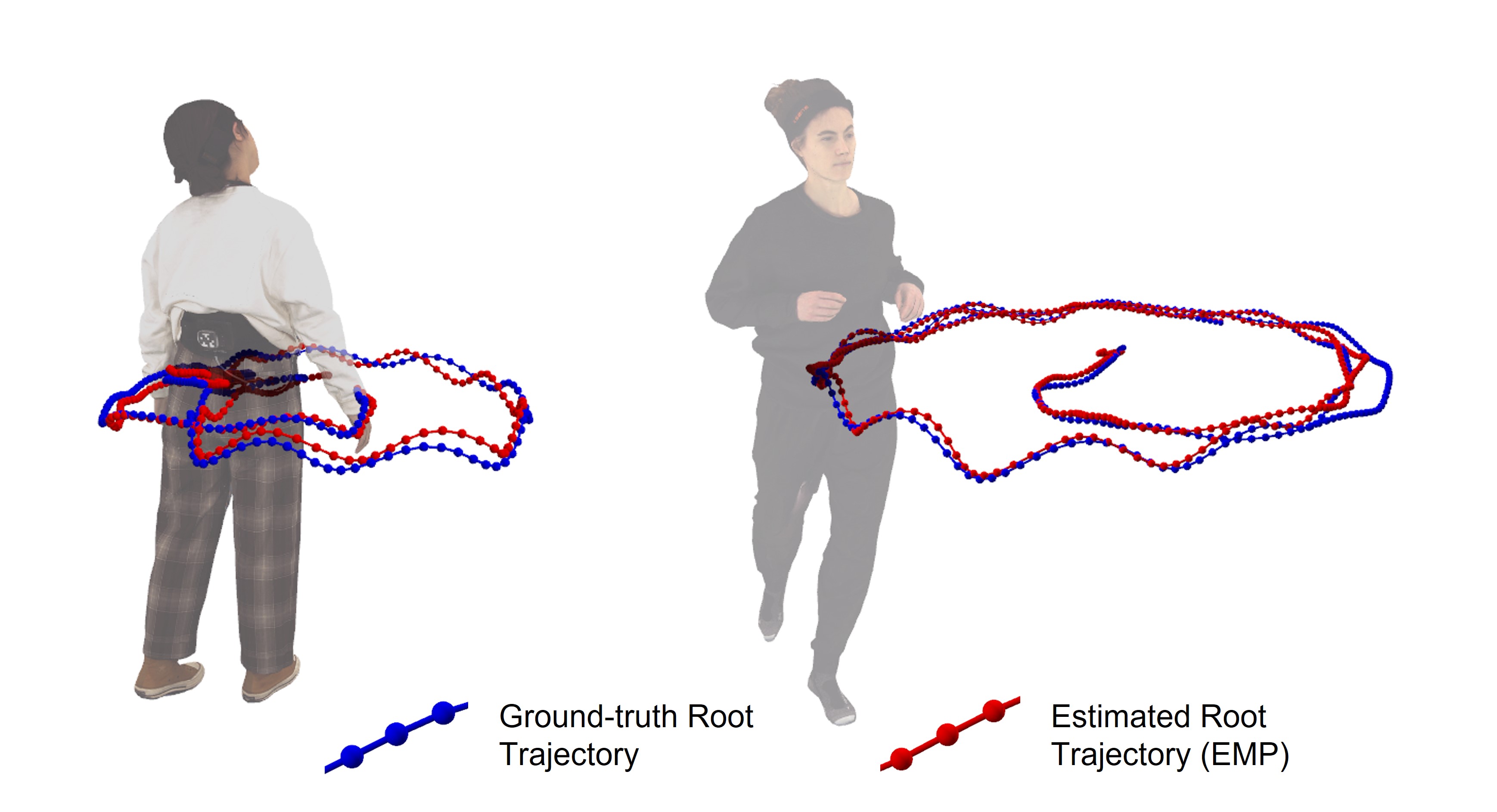}
    \caption{
    Evaluation of global trajectories on our MVS.}
    \label{fig:global_trajectory_alignment}
\end{figure}

\begin{figure}
    \includegraphics[width=\linewidth, trim={0 0 0 0},clip]{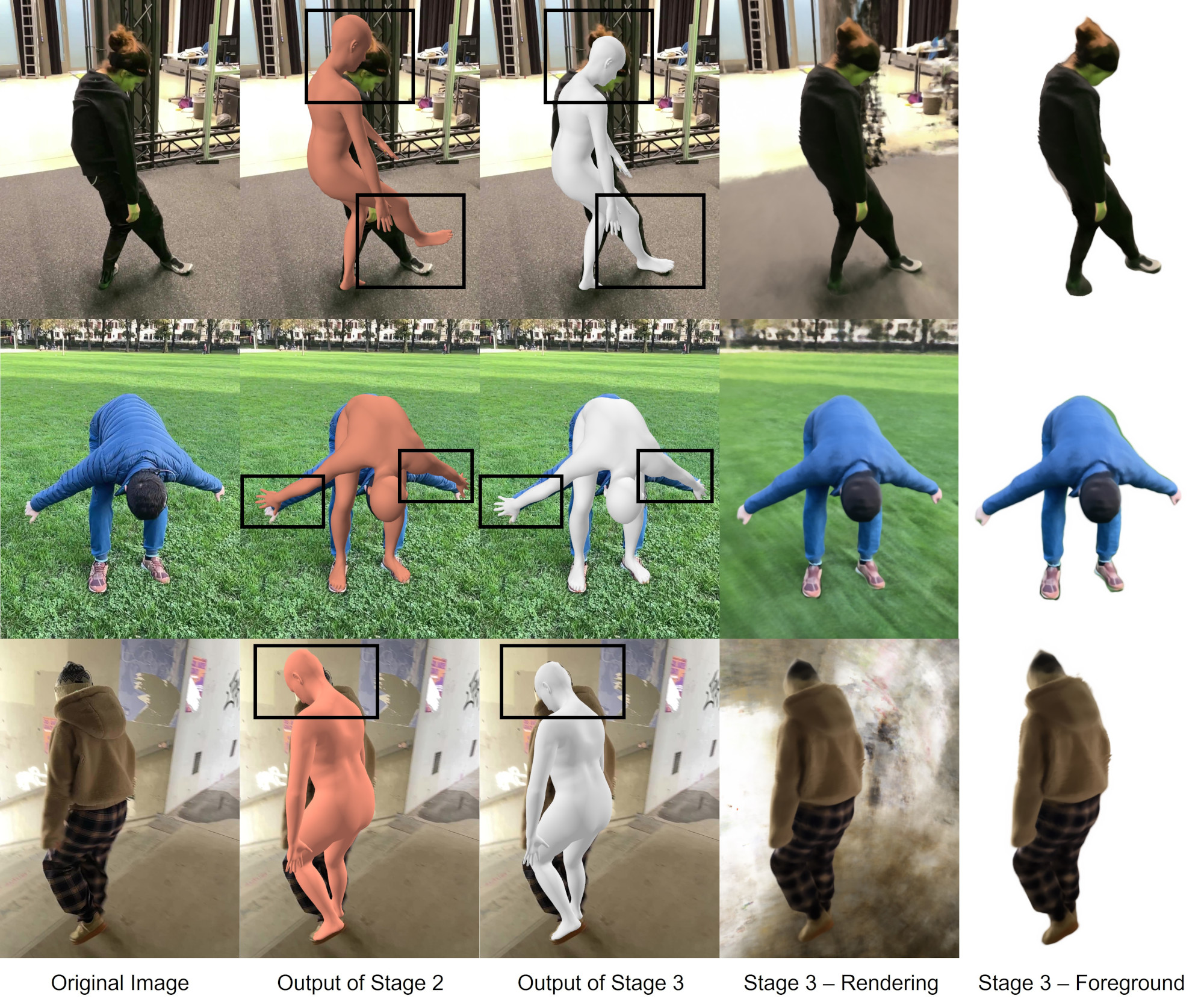}
    \caption{
    Effect of Stage 3. We visualize the output of stage 2 (second column) and the refined output of stage 3 (third column) showing improved pose-to-image alignment. The two right-most columns show the rendering of the entire scene and the separated human (foreground).}
    \label{fig:stage3_effect}
\end{figure}

\subsection{Pose Accuracy}
\label{sec:evaluation_gt}

To estimate the accuracy of \methodname we recorded a number of sequences with the same capture setup as we use for the in-the-wild sequences, but the motions are performed on our MVS \cite{MRCS} that is synchronized with the EM sensors and the iPhone. We use the surface scans and 53 high-resolution RGB views from this stage to procure SMPL ground-truth registrations (see \supplementary for details), which we can then compare to the outputs of \methodname to estimate its accuracy.
We have recorded a total of 21 sequences (approx. 13k frames) distributed over all 10 participants for this evaluation. The respective ablation studies and comparisons to other methods are listed in \tabref{tab:evaluation_gt}.

The closest related in-the-wild dataset to ours is 3DPW \cite{vonMarcard20183dpw}. It is also the only other dataset that provides ground-truth evaluations of their method.
As different sensor technologies are used, a direct comparison to their method is not feasible.
Still, to allow for a comparison of the estimated accuracy, we compute and report the same metrics as \cite{vonMarcard20183dpw}, \ie, the Procrustes-aligned mean per-joint positional and angular errors (MPJPE-PA, MPJAE-PA). To measure smoothness, we follow TransPose \cite{Yi2021TransPose} and report their jitter metric. In addition we show qualitative comparisons to 3DPW with similar motions in the \supplementary

\noindent\textbf{Results}: \tabref{tab:evaluation_gt} allows to draw several conclusions.
First, recent monocular methods - whether they use ground-truth bounding boxes (HybrIK \cite{Li2021hybrik}) or not (ROMP \cite{Sun2021ROMP}) - are far below \methodname's accuracy. Also V2A \cite{Guo2023v2a} suffers without good initial poses. LGD \cite{song2020human}, which uses 2D keypoints in a hybrid optimization and outperforms SPIN \cite{kolotouros2019learning} and Simplify \cite{bogo2016keep} on 3DPW, underperforms compared to \methodname. This highlights a clear need for sensor-based methods to procure high-quality 3D poses.

Second, \tabref{tab:evaluation_gt} ablates the contributions of the multi-stage design of \methodname. We observe that the first stage, which only fits to the EM measurements, already produces good results. Further, the joint optimization in our second stage finds a good trade-off and even improves the initial poses from the first stage via the addition of $E^*_\text{rec}$ and $E_\text{prior}$.
Lastly, the third stage only improves the pose marginally, but helps with smoothness and image alignment (``after $E_\text{S3}$'' in \tabref{tab:evaluation_gt}).
We perform a light smoothing pass as a post-processing step on the outputs of $E_\text{S3}$. We found that this further reduces jitter without breaking pose-to-image alignment. For a visualization of the effect of stage 3, as well as renderings of the neural implicit human model and the scene, please refer to \figref{fig:stage3_effect}. Note that na\"ively smoothing the outputs of the second stage impacts the alignment negatively, which we show in the \supplementary

\subsection{Global Trajectories}
\label{sec:evaluation_traj}

\paragraph*{iPhone Pose Accuracy}
We first compare the iPhone's self-localized poses using optical tracking with our MVS. To do so we rigidly attach an Apriltag \cite{olson2011apriltag1,wang2016apriltag2,krogius2019apriltag3} to the iPhone and move the pair around.
An Apriltag of roughly 5~cm side length can be tracked with millimeter accuracy.
To compare its pose to the iPhone's pose, we must compute an alignment, the details of which are reported in the \supplementary
After alignment, the difference between the iPhone and Apriltag trajectories on a 15 second sequence is $1.8 \pm 0.9$ cm and $0.4 \pm 0.2$ deg respectively.

\paragraph{Global SMPL Trajectories}
To evaluate the accuracy of the global trajectories, we asked half of our participants to move freely in the capture space while we track the iPhone with an Apriltag as above.
This enables us to align the iPhone's and the MVS' tracking frames. For details, please refer to the \supplementary
After alignment, we compute the Euclidean distance between \methodname's predicted trajectory and the ground-truth trajectory obtained on the stage. Over 5 sequences (approx. 3.9k frames) we found that \methodname's trajectories are on average $5.1 \pm 3.2$ cm close to the ground-truth, which is low considering a capture space diameter of $2.5$~meters (see \figref{fig:global_trajectory_alignment} for a visualization).

To gauge the accuracy of the global trajectories in-the-wild, where we cannot track the iPhone, we asked some participants to return to the starting point at the end of the sequence.
This allows us to compute a measure of drift for the in-the-wild sequences.
For an indoor sequence of $81$ meters, this error is $23.4$ cm (or $0.3\%$ of the total path length) and for an outdoor sequence of $112$ meters length it is $73.0$ cm ($0.7\%$) respectively (see also \figref{fig:visualization_glamr} for a visualization).

\section{EMDB}
\label{sec:emdb}

\begin{table}
\centering
\resizebox{1.00\linewidth}{!}{
\begin{tabular}{lccccccc}
\toprule
\multicolumn{1}{c}{Dataset} & \multicolumn{2}{c}{\# number of:} & Size  & \multicolumn{2}{c}{PA Accuracy} & \multicolumn{1}{c}{Global}\\
\cline{2-3}
\cline{5-6}
 &
  \multicolumn{1}{c}{subj.} &
  \multicolumn{1}{c}{seqs.} &
  \multicolumn{1}{c}{[min.]} &
  \multicolumn{1}{c}{MPJPE} &
  \multicolumn{1}{c}{MPJAE} &
  \multicolumn{1}{c}{Traj.}
  \\ \toprule
3DPW \cite{vonMarcard20183dpw} & 7 & 60 & 29.3 & 2.6 cm & 12.1 \textdegree & \xmark \\
\datasetname (Ours) & \textbf{10} & \textbf{81} & \textbf{58.3} & \textbf{2.3} cm & \textbf{10.6} \textdegree & \cmark\\
\bottomrule
\end{tabular}
}
\caption{
Comparison to in-the-wild datasets that provide evaluations of their accuracy. PA: Procrustes-aligned.}
\label{tab:compare_datasets}
\end{table}

\begin{figure}
    \includegraphics[width=\linewidth, trim={0 0 0 0},clip]{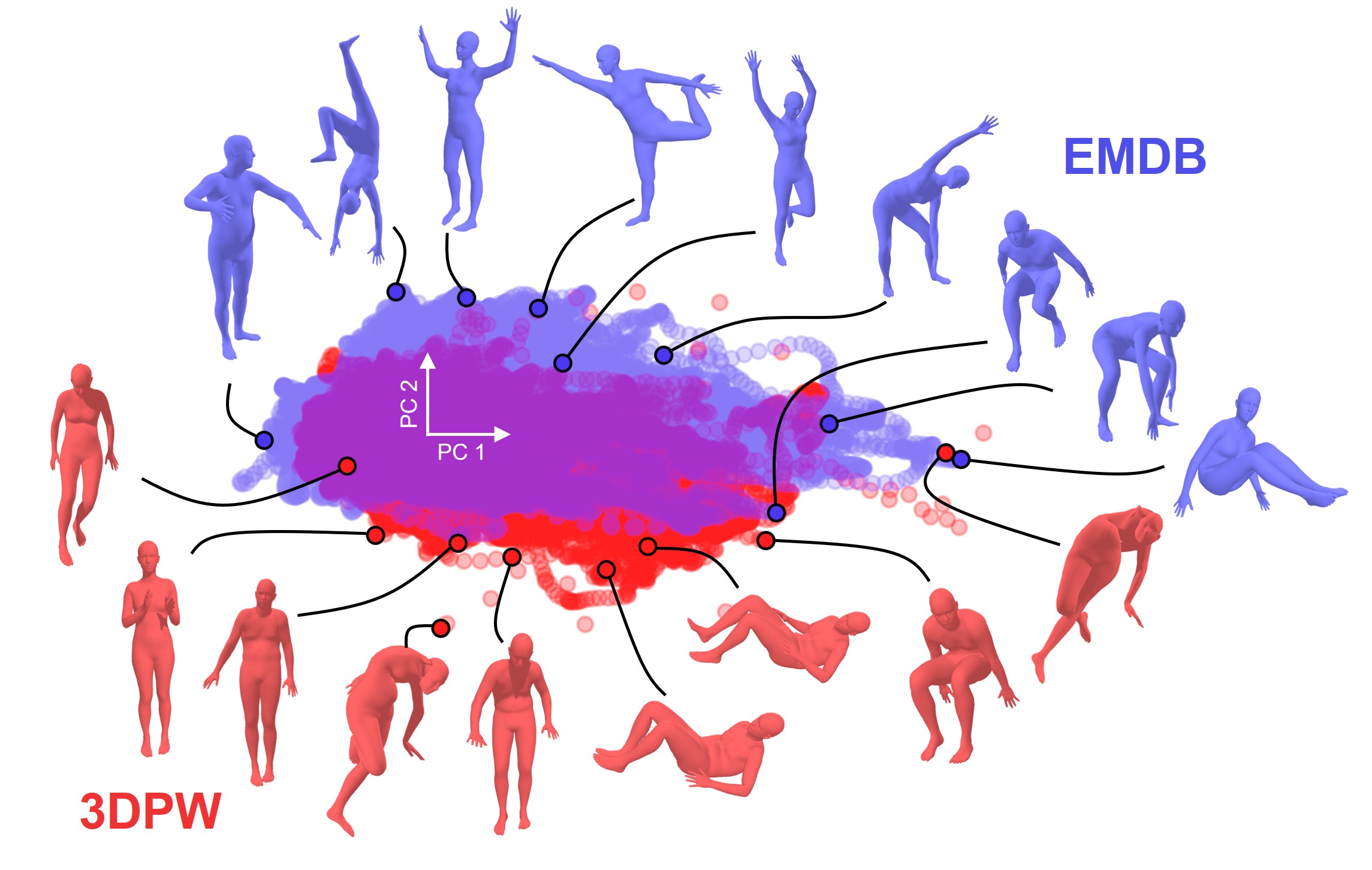}
    \caption{
    Scatter plot of first two principal components computed on 3DPW and EMDB in VPoser's \cite{pavlakos2019expressive} latent space and associated 3D poses for selected data points.}
    \label{fig:pca_analysis}
\end{figure}

\begin{figure}
    \includegraphics[width=\linewidth, trim={0 0 0 0},clip]{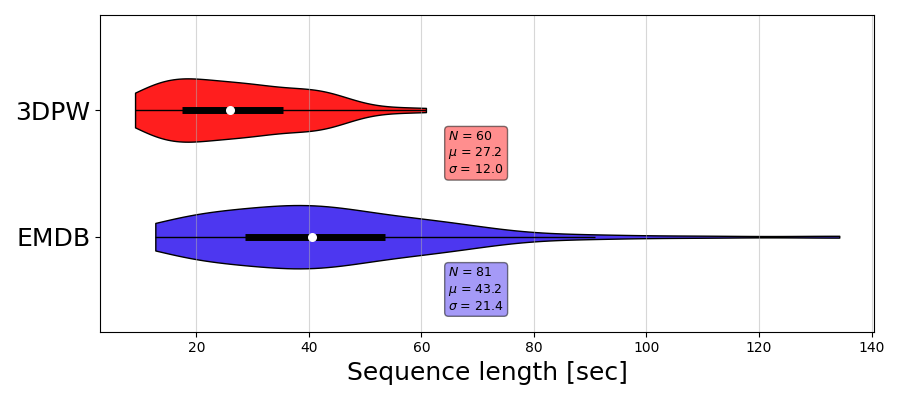}
    \caption{
    Distribution of sequence lengths in seconds in \datasetname and 3DPW (thicker line from $1^{st}$ to $3^{rd}$ quartile).}
    \label{fig:seq_lengths}
\end{figure}

\begin{figure}
    \includegraphics[width=\linewidth, trim={0 0 0 0},clip]{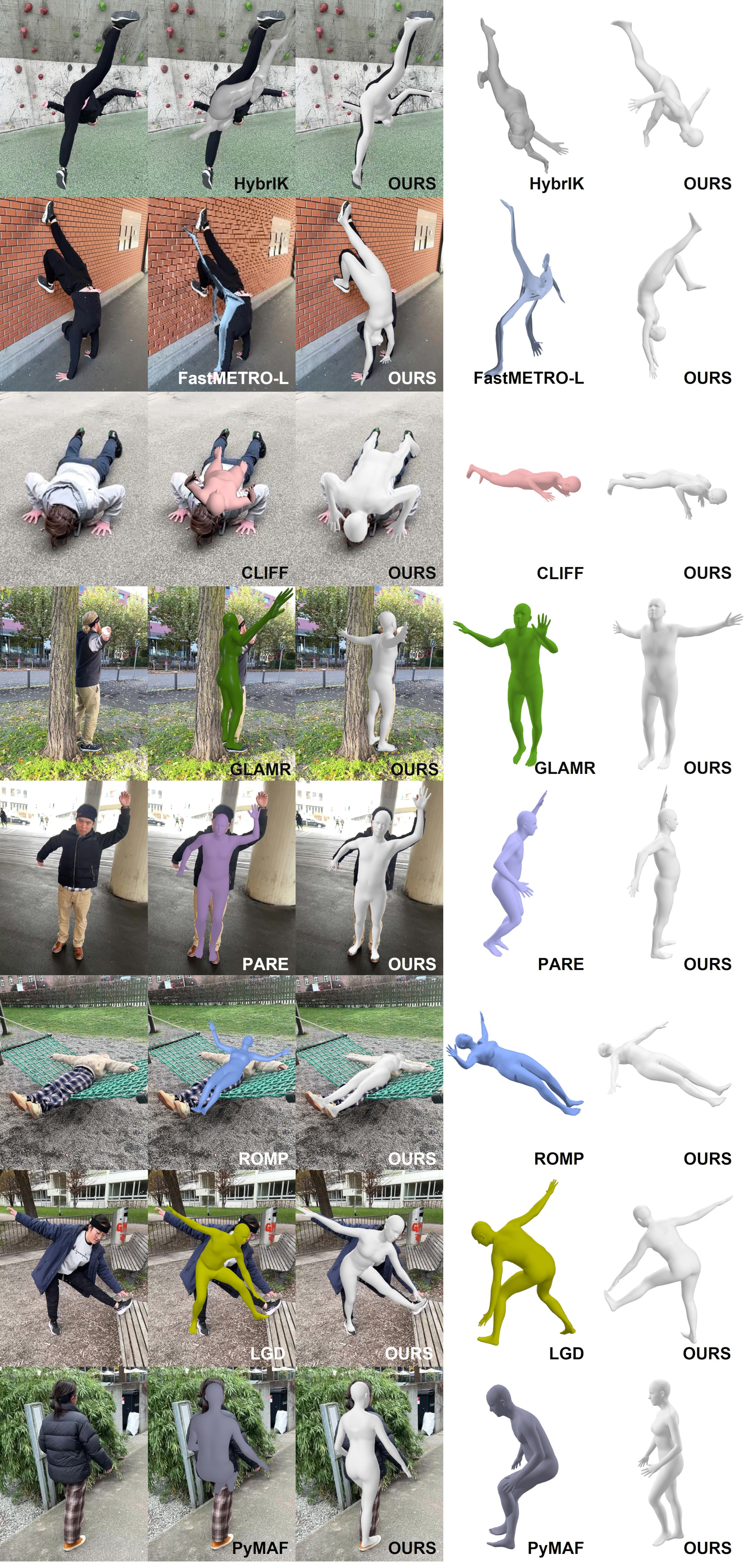}
    \caption{
    Example images and reference poses appearing in \datasetname, alongside comparisons to the outputs of recent state-of-the-art RGB-based pose estimation methods.}
    \label{fig:visualizations}
    \vspace{-0.2cm}
\end{figure}

\definecolor{LightGray}{gray}{0.9}

\begin{table*}
\centering
\resizebox{1.00\linewidth}{!}{
\begin{tabular}{lcc|cc|cc|c}
\toprule
Method & MPJPE $\downarrow$ & MPJPE-PA  $\downarrow$ & MVE  $\downarrow$ & MVE-PA  $\downarrow$ & MPJAE  $\downarrow$ & MPJAE-PA  $\downarrow$ & Jitter  $\downarrow$ \\
       & [mm]  & [mm]     & [mm] & [mm]   & [deg] &   [deg]  & [10m s$^{-3}$] \\ \hline
\rowcolor{LightGray}
PyMAF \cite{pymaf2021} & 131.1 $\pm$ 54.9 & 82.9 $\pm$ 38.2 & 160.0 $\pm$ 64.5 & 98.1 $\pm$ 44.4 & 28.5 $\pm$ 12.5 & 25.7 $\pm$ 10.1 & 81.8 $\pm$ 25.6\\
LGD \cite{song2020human} & 115.8 $\pm$ 64.5 & 81.1 $\pm$ 51.1 & 140.6 $\pm$ 75.8 & 95.7 $\pm$ 56.8 & 25.2 $\pm$ 13.3 & 25.6 $\pm$ 15.3 & 73.0 $\pm$ 38.5\\
\rowcolor{LightGray}
ROMP \cite{Sun2021ROMP} & 112.7 $\pm$ 48.0 & 75.2 $\pm$ \textbf{33.0} & 134.9 $\pm$ 56.1 & 90.6 $\pm$ \underline{38.4} & 26.6 $\pm$ 10.4 & 24.0 $\pm$ \underline{8.7} & 71.3 $\pm$ 25.3\\
PARE \cite{Kocabas2021PARE} & 113.9 $\pm$ 49.5 & 72.2 $\pm$ 33.9 & 133.2 $\pm$ 57.4 & 85.4 $\pm$ 39.1 & 24.7 $\pm$ \textbf{9.8} & \underline{22.4} $\pm$ 8.8 & 75.1 $\pm$ 22.5\\
\rowcolor{LightGray}
GLAMR \cite{yuan2022glamr} & 107.8 $\pm$ 50.1 & 71.0 $\pm$ 36.6 & 128.2 $\pm$ 58.5 & 85.5 $\pm$ 40.9 & 25.5 $\pm$ 12.6 & 23.5 $\pm$ 11.4 & 67.4 $\pm$ 32.3\\
FastMETRO-L \cite{Cho2022FastMETRO} & 115.0 $\pm$ 95.1 & 72.7 $\pm$ 47.4 & 133.6 $\pm$ 109.7 & 86.0 $\pm$ 55.4 & 25.1 $\pm$ 16.0 & 22.9 $\pm$ 12.7 & 81.3 $\pm$ 38.7\\
\rowcolor{LightGray}
CLIFF \cite{li2022cliff} & \underline{103.1} $\pm$ \textbf{43.7} & 68.8 $\pm$ 33.8 & 122.9 $\pm$ \textbf{49.5} & 81.3 $\pm$ \textbf{37.9} & \textbf{23.1} $\pm$ \underline{9.9} & \textbf{21.6} $\pm$ \textbf{8.6} & \underline{55.5} $\pm$ \textbf{17.9}\\
FastMETRO-L* \cite{Cho2022FastMETRO} & 108.1 $\pm$ 52.9 & \underline{66.8} $\pm$ 36.6 & \textbf{119.2} $\pm$ 59.7 & \underline{81.2} $\pm$ 43.9 & n/a & n/a & 185.9 $\pm$ 51.0\\
\rowcolor{LightGray}
HybrIK \cite{Li2021hybrik} & \textbf{103.0} $\pm$ \underline{44.3} & \textbf{65.6} $\pm$ \underline{33.3} & \underline{122.2} $\pm$ \underline{50.5} & \textbf{80.4} $\pm$ 39.1 & \underline{24.5} $\pm$ 11.3 & 23.1 $\pm$ 11.1 & \textbf{49.2} $\pm$ \underline{18.5}\\
\bottomrule
\end{tabular}
}
\caption{
Evaluations of state-of-the-art methods on \datasetname1. Ordered descendingly by MPJPE-PA. Best results in \textbf{bold}, second best \underline{underlined}. FastMETRO-L*: version without SMPL regression head, \ie, the MPJPE is only evaluated on 14 joints as dictated by its model architecture.}
\label{tab:evaluation_baselines}
\end{table*}

\subsection{Dataset Overview}
EMDB contains $10$ participants ($5$ female, $5$ male), who were recorded in a total of $81$ sequences at $30$ fps, resulting in $104,963$ frames or $58.3$ minutes of motion data. We plot the distribution of sequence lengths in \figref{fig:seq_lengths}. The ethnic distribution of participants in \datasetname is: Middle Eastern (1), Asian (3), Caucasian (6).
For a summary of statistics and comparison to other in-the-wild datasets that provide evaluations, please refer to \tabref{tab:compare_datasets}.
Of the 105k frames contained in \datasetname, approx.~$85 \%$ are recorded in-the-wild (indoors or outdoors) and the rest were recorded on our MVS. Please refer to the \supplementary for detailed descriptions of every sequence as well as the distribution of body shapes.

Further, to shed more light onto pose diversity of \datasetname compared to our closest related work, 3DPW \cite{vonMarcard20183dpw}, we project all poses of both datasets into VPoser's \cite{pavlakos2019expressive} latent space, run PCA and plot the first two principal components in \figref{fig:pca_analysis}. We make several observations:
\begin{inparaenum}[i)]
  \item \datasetname covers a larger area than 3DPW.
  \item The additional area is made up of complex and diverse poses.
  \item The highlighted poses of 3DPW around the lower boundary lack diversity.
  \item Outliers on 3DPW can be broken poses, while the closest \datasetname~pose is still valid (see right-most pose pair).
\end{inparaenum}

We provide visualizations of our dataset's quality in \figref{fig:visualizations}.
The recording of this dataset has been approved by our institution's ethics committee. All subjects have participated voluntarily and gave written consent for the capture and the release of their data.

\subsection{Baselines on EMDB}
We evaluate two tasks on \datasetname: camera-local 3D human pose estimation from monocular RGB images and the emerging task of global trajectory prediction. To this end we partition \datasetname into two parts: \datasetname1, which consists of our most challenging sequences ($17$ sequences of a total of $24\,117$ frames), and \datasetname2 with $25$ sequences ($43\,120$ frames) featuring meaningful global trajectories.

\paragraph*{Monocular RGB-based Pose Estimation}
We evaluate a total of $8$ recent SOTA methods on \datasetname1.
Please refer to \tabref{tab:evaluation_baselines} for an overview of the results.
We follow the AGORA protocol \cite{Patel2021AGORA} and compute the MPJPE and MVE metrics with both a Procrustes alignment (*-PA) and a hip-alignment pre-processing step.
In addition, we follow sensor-based pose estimation work and report the joint angular error MPJAE and the jitter metric \cite{Yi2021TransPose}.

To provide a fair evaluation and comparison between baselines, we provide ground-truth bounding boxes for methods that accept them or tightly crop the image to the human and re-scale it to the resolution the method requires. Hence only ROMP \cite{Sun2021ROMP} takes the input images as is. Also, we exclude the few frames where the human is entirely occluded.
We use the HRNet version of HybrIK \cite{Li2021hybrik} -- an improved variant of their originally published model.
For FastMETRO \cite{Cho2022FastMETRO} we use their biggest model (*-L) and evaluate both with and without the SMPL regression head.
None of the methods are fed any knowledge about the camera and comparisons to the ground-truth are performed in camera-relative coordinates.
We use the SMPL gender(s) that the respective method was trained with.

\noindent\textbf{Results}: \tabref{tab:evaluation_baselines} reveals HybrIK \cite{Li2021hybrik} as the best performer. Nonetheless, an MPJPE-PA error of $> 65$ mm suggests that there is a lot of room for improvement.
As is noted in AGORA \cite{Patel2021AGORA}, we highlight that the MPJPE-PA is a very forgiving metric due to the Procrustes alignment that removes rotation, translation, and scale. We have noticed that a good MPJPE-PA does not always translate to visually pleasing results, a circumstance that the rather high jitter and MPJPE value for all baselines supports  (see also the supp. video).
Similarly we observe very high standard deviations, which is a metric that tends to have been neglected by common benchmarks.
Furthermore, we notice high angular errors of $> 23^{\circ}$ on average for all methods.
These results and the fact that we used ground-truth bounding-boxes for all methods except ROMP, suggest that there is ample space for future research in this direction using \datasetname.

\begin{figure}
    \includegraphics[width=\linewidth, trim={0 0 0 0},clip]{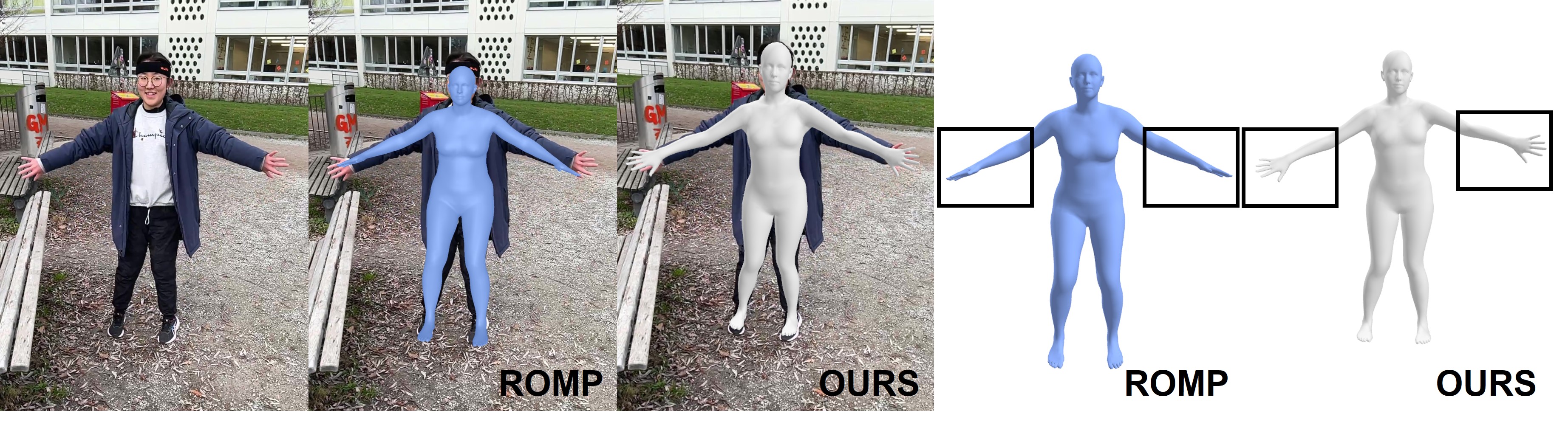}
    \caption{Common failure case where the baseline (here ROMP \cite{Sun2021ROMP}) fails to capture the lower arm rotations.}
    \label{fig:visualize_lower_arms}
\end{figure}

We show selected results for each baseline in \figref{fig:visualizations} and further highlight a common failure case in \figref{fig:visualize_lower_arms} where the baseline method fails to capture the lower arm rotations. Note that such a failure case is not accounted for by the MPJPE metric, which is why we also report angular errors.

\begin{figure}
    \includegraphics[width=\linewidth, trim={0 0 0 0},clip]{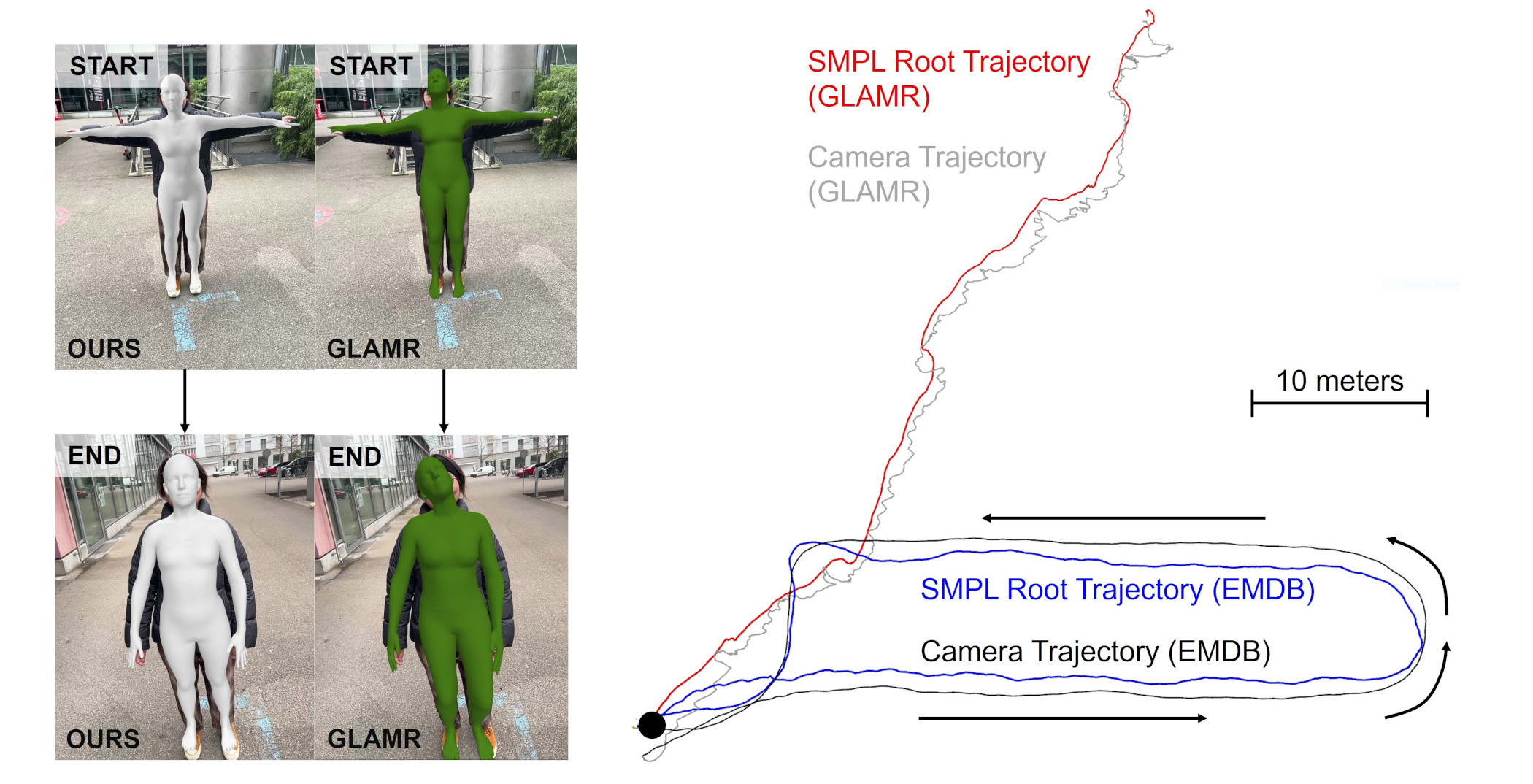}
    \caption{
    (Left) GLAMR \cite{yuan2022glamr} results projected into the camera at the start and end of a loop-closing sequence. (Right) GLAMR's global trajectories compared to ours.}
    \label{fig:visualization_glamr}
\end{figure}

\paragraph*{Global Trajectory Estimation}
As a second task, we evaluate GLAMR \cite{yuan2022glamr} on \datasetname2.
We use GLAMR's publicly available code to run and evaluate its performance.
This protocol computes global MPJPE, MVE, and acceleration metrics on windows of 10 seconds length, where the beginning of each window is aligned to the ground-truth trajectory.
We found that GLAMR achieves a G-MPJPE of $3\,193$~mm, a G-MVE of $3\,203$~mm and acceleration of $12.6$~mm~s$^{-2}$.
We visualize one sequence in \figref{fig:visualization_glamr}, where we observe that the GLAMR prediction drifts significantly from our provided trajectories.
We believe \datasetname will help to boost future method's performance on this task.
\section{Conclusion}

\noindent\textbf{Conclusion} \
We present \datasetname, the first comprehensive dataset to provide accurate SMPL poses, shapes and trajectories in an unrestricted, mobile, in-the-wild setting.
Our results indicate a clear need for sensor-based performance capture to procure high-quality 3D human motion and push the boundaries of monocular RGB-based pose estimators.

\noindent \textbf{Limitations} \
\datasetname does not contain multi-person sequences, because using multiple EM systems requires non-trivial changes to avoid cross-talk and interference between sensors.
Furthermore, there are no sensors on the feet as indoor floors often contain metal beams that would disturb the readings.
Lastly, the quality of our camera trajectories is upper-bounded by the quality of Apple's AR toolkit.

\noindent \textbf{Acknowledgments} \ 
We thank Robert Wang, Emre Aksan, Braden Copple, Kevin Harris, Mishael Herrmann, Mark Hogan, Stephen Olsen, Lingling Tao, Christopher Twigg, and Yi Zhao for their support. Thanks to Dean Bakker, Andrew Searle, and Stefan Walter for their help with our infrastructure. Thanks to Marek, developer of record3d, for his help with the app. Thanks to Laura Wülfroth and Deniz Yildiz for their assistance with capture. Thanks to Dario Mylonopoulos for his priceless work on aitviewer which we used extensively in this work. We are grateful to all our participants for their valued contribution to this research. Computations were carried out in part on the ETH Euler cluster.

{\small
\bibliographystyle{ieee_fullname}
\bibliography{egbib}
}

\clearpage
\newpage
\renewcommand\thesection{\Alph{section}}
\setcounter{section}{0}
\section{Sensor Placement}
\begin{figure}[b]
    \includegraphics[width=\linewidth, trim={0 0 0 0},clip]{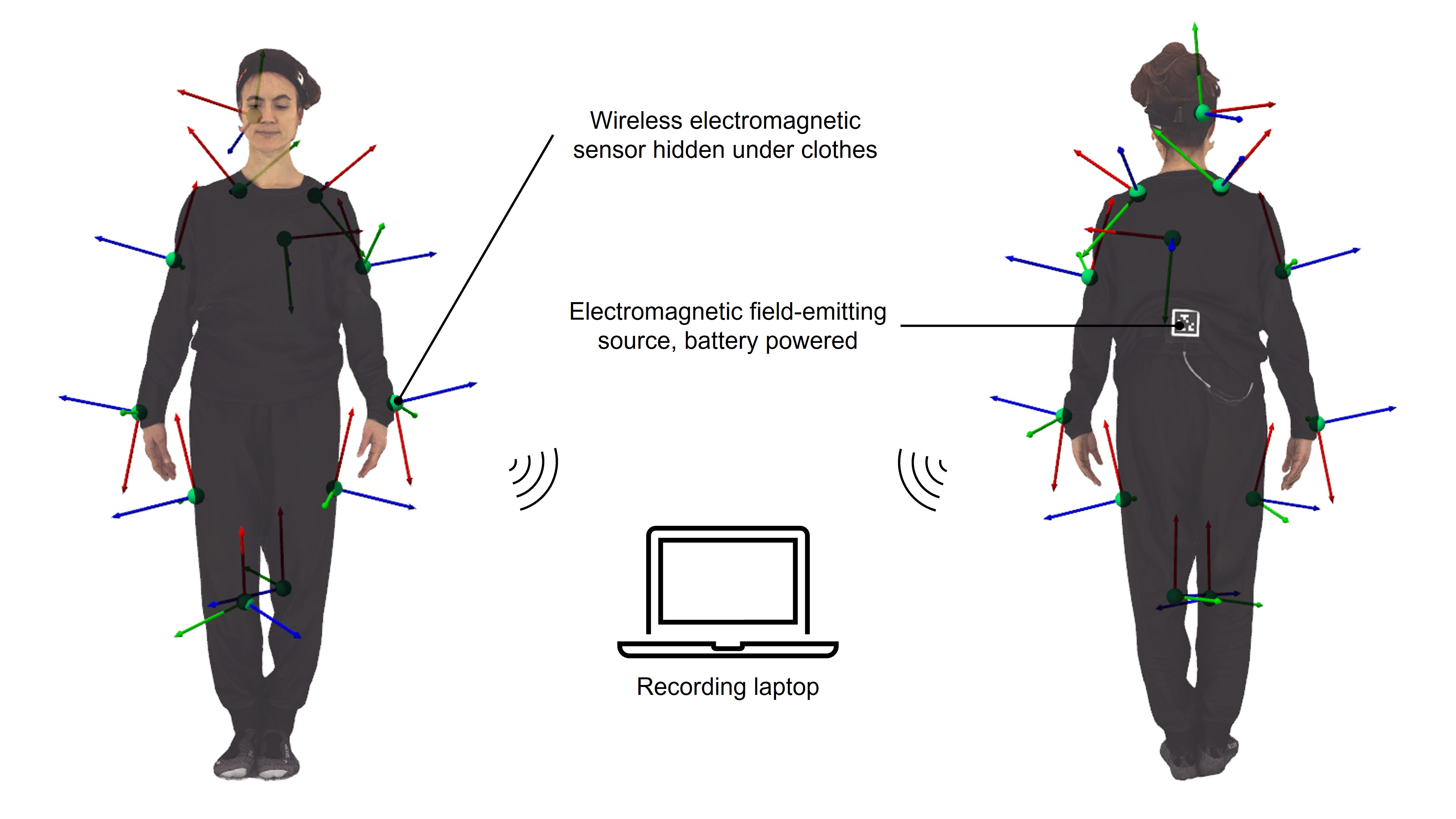}
    \caption{Sensor placement and recording setup. The Apriltag on the source is only required for the body calibration.}
    \label{fig:sensor_placement}
\end{figure}

We place sensors under regular clothing as shown in \figref{fig:sensor_placement}. All sensors communicate wirelessly with two receivers plugged into a recording laptop via USB. The laptop usually stays within 3-4 meters of the participant to minimize packet loss. The sensors measure their position and orientation relative to the EM field emitting source mounted on the lower back. For more details please refer to \cite{kaufmann2021empose}. As seen in \figref{fig:sensor_placement} an Apriltag \cite{olson2011apriltag1,wang2016apriltag2,krogius2019apriltag3} is attached to the source, which we only require for the body calibration as explained in more detail in \secref{sec:supp_body_calibration}. Sensors and source are battery-powered, which can all be neatly stowed away under regular clothing.
For a depiction of how the sensors are strapped to the body, please refer to \figref{fig:em_calibration}.

\begin{figure}[b]
    \includegraphics[width=\linewidth, trim={0 0 0 0},clip]{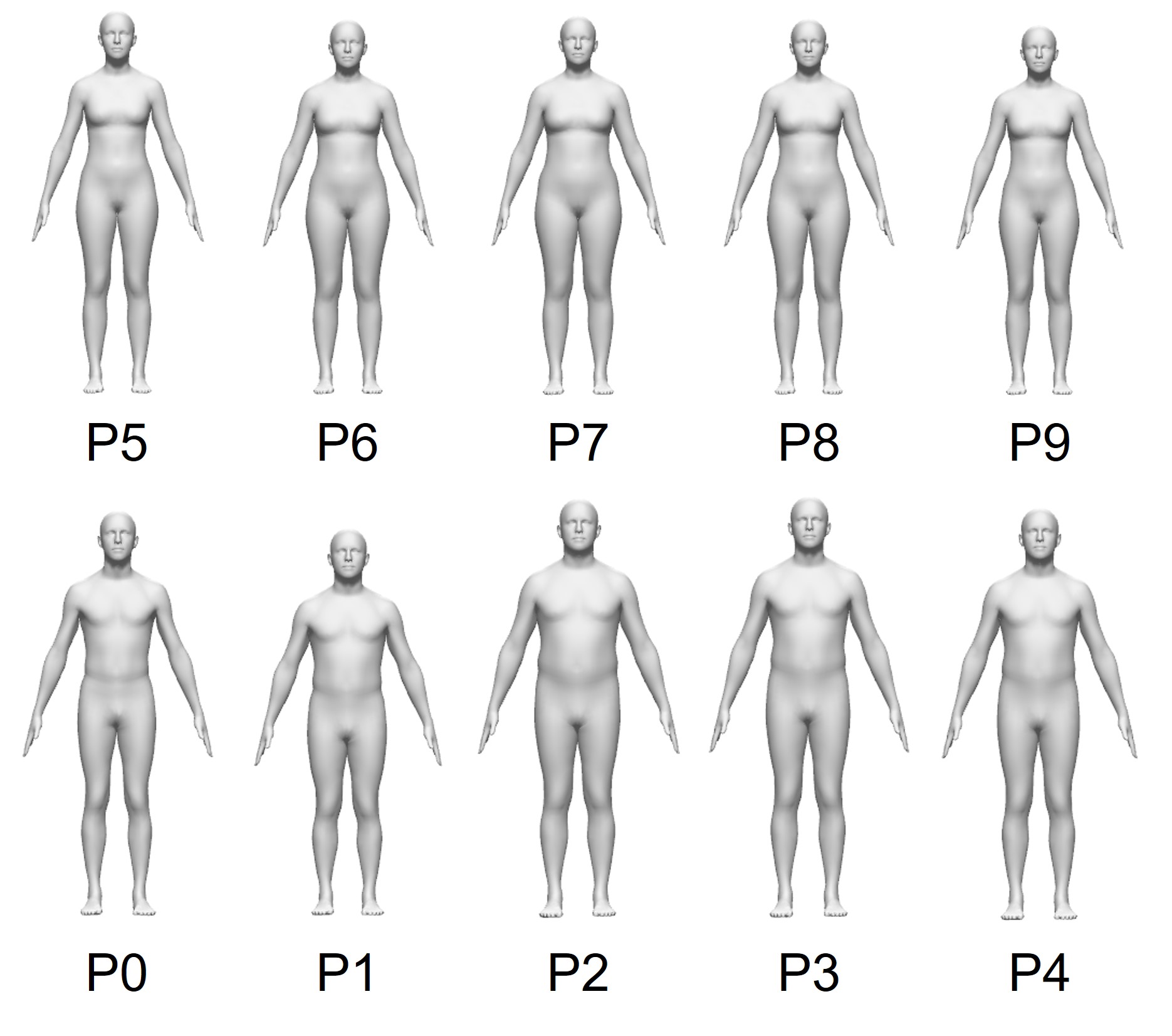}
    \caption{Shapes of all 10 participants appearing in \datasetname. For reference, the height of P0 is 177 cm.}
    \label{fig:shape_distribution}
\end{figure}

\section{Details on \datasetname's Contents}

\renewcommand{\cellalign}{tl}
\begin{table}
\centering
\resizebox{1.00\linewidth}{!}{
\begin{tabular}{cccp{4.0cm}c}
\toprule
 \textbf{Subj.} & \textbf{Seq.} & \multirow{2}{*}{\textbf{Loc.}} & \multirow{2}{*}{\textbf{Activity}} & \multirow{2}{*}{\textbf{\# Frames}} \\
 \textbf{ID}    & \textbf{ID}  & & & \\
 \toprule
 P0 & 0 & MVS & \makecell{arm rotation, leg raises,\\ jump} & $381$\\
 P0 & 1 & MVS & \makecell{walk, crouch, bend,\\ arm swing, arm raises} & $554$\\
 P0 & 2 & MVS & punches, pirouette, leg curls & $543$\\
 P0 & 3 & MVS & upper body range of motion& $661$\\
 P0 & 4 & MVS & upper body range of motion& $667$\\
 P0 & 5 & MVS & kicks, punches, leg raises, arm swings & $635$\\
 P0 & 6 & I & jumping, swirl, boxing & $858$\\
 P0 & 7 & O & lunges, push-ups & $1\,277$\\
 P0 & 8 & O & remove jacket & $591$\\
 P0 & 9 & O & long walk, straight line & $2\,009$\\
 P0 & 10 & O & range of motion & $1\,183$\\
 \toprule
 P1 & 11 & MVS & range of motion & $951$\\
 P1 & 12 & MVS & walk in circle, punch, kick, crouch, upper body twist & $991$\\
 P1 & 13 & O & very long walk & $4\,028$\\
 P1 & 14 & O & \makecell{climb on platform, sit, \\ jog around platform} & $1\,284$\\
 P1 & 15 & O & range of motion & $1\,348$\\
 P1 & 16 & O & \makecell{warm-up, side-stepping, \\ push-ups} & $1\,365$\\
 \toprule
 P2 & 17 & MVS & arm and leg motions, jump & $574$\\
 P2 & 18 & MVS & \makecell{occluded arm motions,\\ crouch}  & $577$\\
 P2 & 19 & I & walk off stage & $1\,299$\\
 P2 & 20 & O & long walk & $2\,713$\\
 P2 & 21 & O & sit, stand and balance & $1\,272$\\
 P2 & 22 & O & play with basketball & $1\,438$\\
 P2 & 23 & O & hug tree & $1\,086$\\
 P2 & 24 & O & long walk, climbing & $3\,280$\\
 \toprule
 P3 & 25 & MVS & walk while moving arms & $528$\\
 P3 & 26 & MVS & \makecell{cross arms, arm motions\\ with occlusions, squats} & $557$\\
 P3 & 27 & I/O & walk off stage & $1\,448$\\
 P3 & 28 & O & lunges while walking & $1\,836$\\
 P3 & 29 & O & walk up stairs & $1\,205$\\
 P3 & 30 & O & walk down stairs & $1\,128$\\
 P3 & 31 & O & workout & $1\,216$\\
 P3 & 32 & O & soccer warmup 1 & $1\,084$\\
 P3 & 33 & O & soccer warmup 2 & $1\,433$\\
 \toprule
 P4 & 34 & MVS & walk, upper body twist, arm motions & $541$\\
 P4 & 35 & I & walk along hallway & $1\,226$\\
 P4 & 36 & O & long walk & $2\,160$\\
 P4 & 37 & O & jog in circle & $881$\\
 \toprule
 \multicolumn{3}{l}{\textbf{Total}} & & $\mathbf{46\,808}$\\
 \bottomrule
\end{tabular}
}
\caption{More detailed description of the sequences appearing in \datasetname for subjects P0 to P4 (male participants). For subjects P5 to P9 please refer to table \tabref{tab:emdb_content_2}. \textit{Loc.} refers to where the recording took place (\textit{MVS}: In our multi-view volumetric capture studio, \textit{I}: Indoor, \textit{O}: Outdoor). The data is recorded at 30 fps (26 minutes).}
\label{tab:emdb_content_1}
\end{table}
\renewcommand{\cellalign}{tl}
\begin{table}
\centering
\resizebox{1.00\linewidth}{!}{
\begin{tabular}{cccp{4.0cm}c}
\toprule
 \textbf{Subj.} & \textbf{Seq.} & \multirow{2}{*}{\textbf{Loc.}} & \multirow{2}{*}{\textbf{Activity}} & \multirow{2}{*}{\textbf{\# Frames}} \\
 \textbf{ID}    & \textbf{ID}  & & & \\
 \toprule
 P5 & 38 & MVS & arm and leg motions & $500$\\
 P5 & 39 & MVS & walk and jog in circle & $600$\\
 P5 & 40 & I & walk in big circle & $2\,661$\\
 P5 & 41 & I & jog in circle, workout & $1\,762$\\
 P5 & 42 & I & freestyle dancing & $1\,291$\\
 P5 & 43 & I & drink water & $1\,400$\\
 P5 & 44 & I & range of motion & $1\,381$\\
 \toprule
 P6 & 45 & MVS & range of motion & $994$\\
 P6 & 46 & MVS & \makecell{jumping jacks, lunges,\\ squats, torso twists} & $1\,005$\\
 P6 & 47 & O & slalom with occlusions & $677$\\
 P6 & 48 & O & walk down slope & $1\,959$\\
 P6 & 49 & O & walk down and up big stairs & $1\,559$\\
 P6 & 50 & O & workout & $1\,532$\\
 P6 & 51 & O & dancing, lunges & $1\,427$\\
 P6 & 52 & O & walk behind low wall & $509$\\
 \toprule
 P7 & 53 & MVS & range of motion, walk & $967$\\
 P7 & 54 & MVS & \makecell{crouching, arm crossing,\\ jump, balance on one leg} & $1\,045$\\
 P7 & 55 & O & long walk & $2\,179$\\
 P7 & 56 & O & walk stairs up and down & $1\,120$\\
 P7 & 57 & O & lie, rock on chair & $1\,558$\\
 P7 & 58 & O & parcours! & $1\,332$\\
 P7 & 59 & O & range of motion & $1\,839$\\
 P7 & 60 & O & \makecell{push-ups, dips,\\ jumping jacks} & $1\,693$\\
 P7 & 61 & O & sit on bench, walk & $1\,914$\\
 \toprule
 P8 & 62 & MVS & freestyle movement & $1\,035$\\
 P8 & 63 & MVS & range of motion with occlusions & $1\,007$\\
 P8 & 64 & O & skateboarding & $1\,704$\\
 P8 & 65 & O & walk straight line & $1\,981$\\
 P8 & 66 & O & range of motion & $1\,808$\\
 P8 & 67 & O & sprint back and forth & $801$\\
 P8 & 68 & O & handstand & $1\,606$\\
 P8 & 69 & O & cartwheel, jump & $656$\\
 \toprule
 P9 & 70 & MVS & range of motion & $1\,045$\\
 P9 & 71 & MVS & jog in circle, head motions & $970$\\
 P9 & 72 & O & jump on bench & $707$\\
 P9 & 73 & O & body scanner motions & $1\,264$\\
 P9 & 74 & O & range of motion & $1\,814$\\
 P9 & 75 & O & slalom around tree & $1\,117$\\
 P9 & 76 & O & sitting & $1\,768$\\
 P9 & 77 & O & walk stairs up & $728$\\
 P9 & 78 & O & walk stairs up and down & $1\,083$\\
 P9 & 79 & O & walk in rectangle & $1\,917$\\
 P9 & 80 & O & walk in big circle & $2\,240$\\
 \toprule
 \multicolumn{3}{l}{\textbf{Total}} & & $\mathbf{58\,155}$\\
 \bottomrule
\end{tabular}
}
\caption{More detailed description of the sequences appearing in \datasetname for subjects P5 to P9 (female participants). For subjects P0 to P4 please refer to table \tabref{tab:emdb_content_2}. \textit{Loc.} refers to where the recording took place (\textit{MVS}: In our multi-view volumetric capture studio, \textit{I}: Indoor, \textit{O}: Outdoor). The data is recorded at 30 fps (32.3 minutes).}
\label{tab:emdb_content_2}
\end{table}

We describe the activity of each sequence appearing in \datasetname in more detail in \tabref{tab:emdb_content_1} and \tabref{tab:emdb_content_2}. From these tables we note that $55.4 \%$ ($44.6 \%$) of all frames in \datasetname are performed by female (male) participants. Furthermore, $15.5 \%$ of all data was recorded on our multi-view volumetric capture system (MVS, \cite{MRCS}), $12.7 \%$ was recorded indoors (but not on the MVS) and the remaining $71.8 \%$ of \datasetname were captured outdoors.

\section{SMPL Registration to Multi-View Data}
\label{sec:supp_smpl_registration}
In this section we explain how we obtain SMPL \cite{SMPL:2015} ground-truth registrations from data recorded with our MVS \cite{MRCS}. We use the same procedure to obtain the ground-truth SMPL shape $\vectr{\beta}$ on minimally clothed scans as mentioned in Sec.~4.2 of the main paper and to register SMPL parameters for the pose accuracy evaluations in Sec.~6.1 of the main paper.

\begin{figure}
    \includegraphics[width=\linewidth, trim={0 0 0 0},clip]{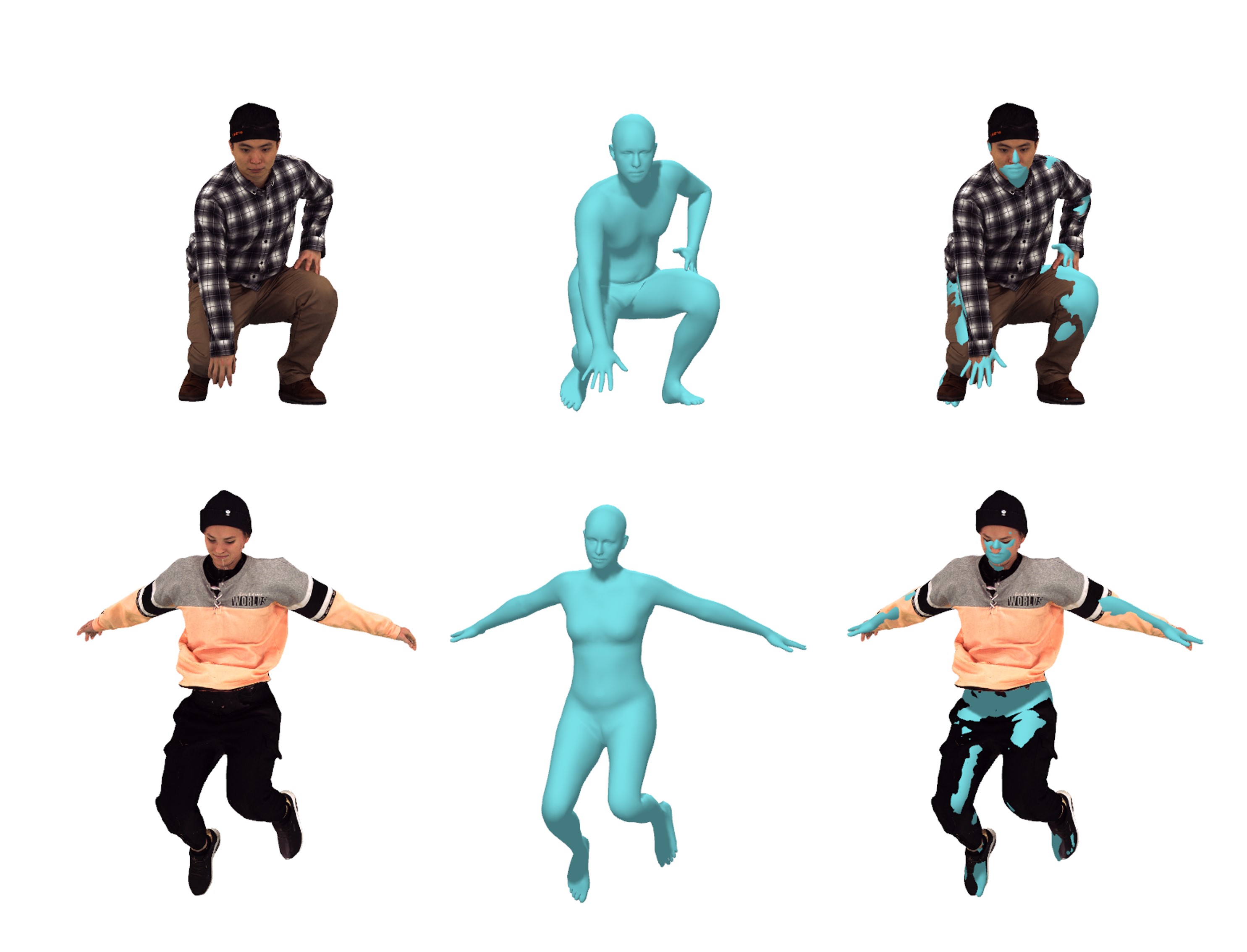}
    \caption{Examples of raw scans (left) and our resulting SMPL registrations (middle). (Right) Scan and SMPL registration overlaid.}
    \label{fig:smpl_registrations}
\end{figure}

Our MVS provides high-quality 3D scans (in the form of watertight meshes with 40k vertices) and high-resolution RGB images from 53 camera views. We fit an SMPL \cite{SMPL:2015} model parameterized by $\vectr{\Omega} = (\vectr{\theta}_r, \vectr{\theta}_b, \mathbf{t}, \vectr{\beta})$ to this data. We use a gender-specific model with the gender that our participants have indicated on a respective questionnaire or a neutral body model if they chose not to answer that question.
For a visualization of scans and registrations, please refer to \figref{fig:smpl_registrations}.

\subsection{3D Keypoint Triangulation}
\label{sec:triangulation}
We start the registration process by detecting Openpose 2D keypoints \cite{Cao2019OpenPoseTPAMI,simon2017OpenPoseHand,wei2016OpenPoseCPM,cao2016realtime} in the multi-view camera images.
When we record sequences with the MVS and the iPhone together, some of the RGB images will show multiple people, \ie, including the person holding the iPhone. This would require running a person-tracker to isolate the correct Openpose Keypoints. To avoid this complication, we instead back-project the high-quality scans obtained from the MVS into the camera views with a white background and then run Openpose on these images.
Note that the quality of the scans is not affected by the presence of a second person on the capture stage, as the assistant holding the iPhone is outside the calibrated capture volume.
Given the 2D keypoint detections in the various views, we triangulate 3D keypoints using ordinary least squares to solve the over-determined linear system. This produces $25$ 3D keypoints per time step in COCO format, denoted as $\mathbf{x}^{3D}_i$.

\subsection{SMPL Fitting}
\label{multi-stage fitting}

After triangulation, we employ an optimization procedure to fit SMPL to the 3D keypoints and scans following \cite{Patel2021AGORA, alldieck2019tex2shape}.
Our implementation follows code published by \cite{bhatnagar2020ipnet,bhatnagar2020loopreg} and uses PyTorch \cite{pytorch}.
We explain the optimization terms and details of the fitting procedure in the following.

\noindent\textbf{3D Keypoint Term} To optimize the pose, a 3D keypoint term is used, where keypoints $\mathbf{\hat{x}}_j^{3D}$ are extracted from the SMPL joints $X(\vectr{\Omega})$ via a pre-defined mapping, and $\mathbf{x}_j^{3D}$ are the triangulated points described in \secref{sec:triangulation}. Joints are weighted via $w_j \in \mathbb{R}$.

\begin{equation}
    E_\text{J} = \frac{1}{J} \sum_{j=1}^J w_j \cdot || \mathbf{x}_j^{3D} -  \mathbf{\hat{x}}_j^{3D} ||_2^2
\end{equation}

\noindent\textbf{Surface Term} To incorporate dense surface information from our scans, we use the following term:

\begin{align}
    E_\text{S} = & \frac{1}{| \mathcal{V}| } \sum_{\mathbf{v} \in \mathcal{V}} \rho (d(\mathbf{v}, \mathcal{M}(\vectr{\Omega}))) +  \nonumber \\
    & \frac{1}{|\mathcal{M}(\vectr{\Omega})|} \sum_{\mathbf{m} \in \mathcal{M}(\vectr{\Omega})} \rho( d (\mathbf{m}, \mathcal{V}))
\end{align}
\noindent where $\mathcal{V}$ is the set of points sampled from the scan, $\mathcal{M}$ the SMPL mesh, $d(\mathbf{p}, \mathcal{R})$ measures the squared Euclidean distance of a point $\mathbf{p} \in \mathbb{R}^3$ to the closest vertex in the point cloud $\mathcal{R}$ and $\rho$ is a generalized robustifier \cite{BarronCVPR2019}. We sample $| \mathcal{V} | = 50\,000$ points on each scan.
To encourage the SMPL mesh to lie within the scan, we follow \cite{alldieck2019tex2shape} and enforce all points $\mathbf{v} \in \mathcal{V}$ that lie outside of the SMPL mesh $\mathcal{M}$ to move inside by increasing their weight in the surface term $E_\text{S}$.

\noindent\textbf{Regularization} It may happen that the SMPL spine is bent unnaturally leading to bulging belly artifacts. To counteract this, we leverage the tracked Apriltag pose on the EM source strapped to the lower back to add a regularizer $E_\text{spine}$. This prior enforces that a set of hand-picked SMPL vertices that are close to the Apriltag remain close to it. We further add regularizers $E_\text{reg}$ to penalize impossible joint angles.

\noindent\textbf{Optimization Details} We use Adam \cite{kingma2014adam} and optimize a given sequence frame-by-frame where we use the previous output as the initialization for the current time step. For every frame, we use two optimization stages. In the first stage we optimize for all parameters $\vectr{\Omega}$ using terms $E_\text{J}, E_\text{S}, E_\text{spine}, E_\text{reg}$. In the second stage we use the same terms, but refine the pose parameters only $(\vectr{\theta}_b, \vectr{\theta}_r)$.

\noindent\textbf{Shape} To deal with shape ambiguity that is caused by loose clothing, AGORA \cite{Patel2021AGORA} uses Graphonomy \cite{Gong2019Graphonomy} to obtain a skin-cloth segmentation. To avoid this rather time-consuming procedure, we instead disentangle the shape from the pose optimization.
To do so, we first scan participants in minimal, tight-fitting clothing while they perform an easy A-pose. We then run the registration pipeline on this sequence to obtain the shape $\vectr{\beta} \in \mathbb{R}^{10}$. Henceforth, the shape is fixed and no longer treated as an optimization parameter.

\section{Body Calibration Details}
\label{sec:supp_body_calibration}
In this section we provide more details on how we calibrate skin-to-sensor offsets for each subject as mentioned in Sec.~4.2 of the main paper.

\subsection{EM and MVS Alignment}
\label{sec:supp_em_alignment}
To compute skin-to-sensor offsets we must first spatially and temporally align the EM space with our MVS. For temporal alignment we use the Atomos Ultrasync One Box \cite{atomosTimecode} to generate a timecode that we feed via LTC to our MVS and the EM sensors. Both the cameras and the EM sensors can be triggered via LTC timecode allowing for a precise temporal alignment.

\begin{figure}
    \includegraphics[width=\linewidth, trim={0 0 0 0},clip]{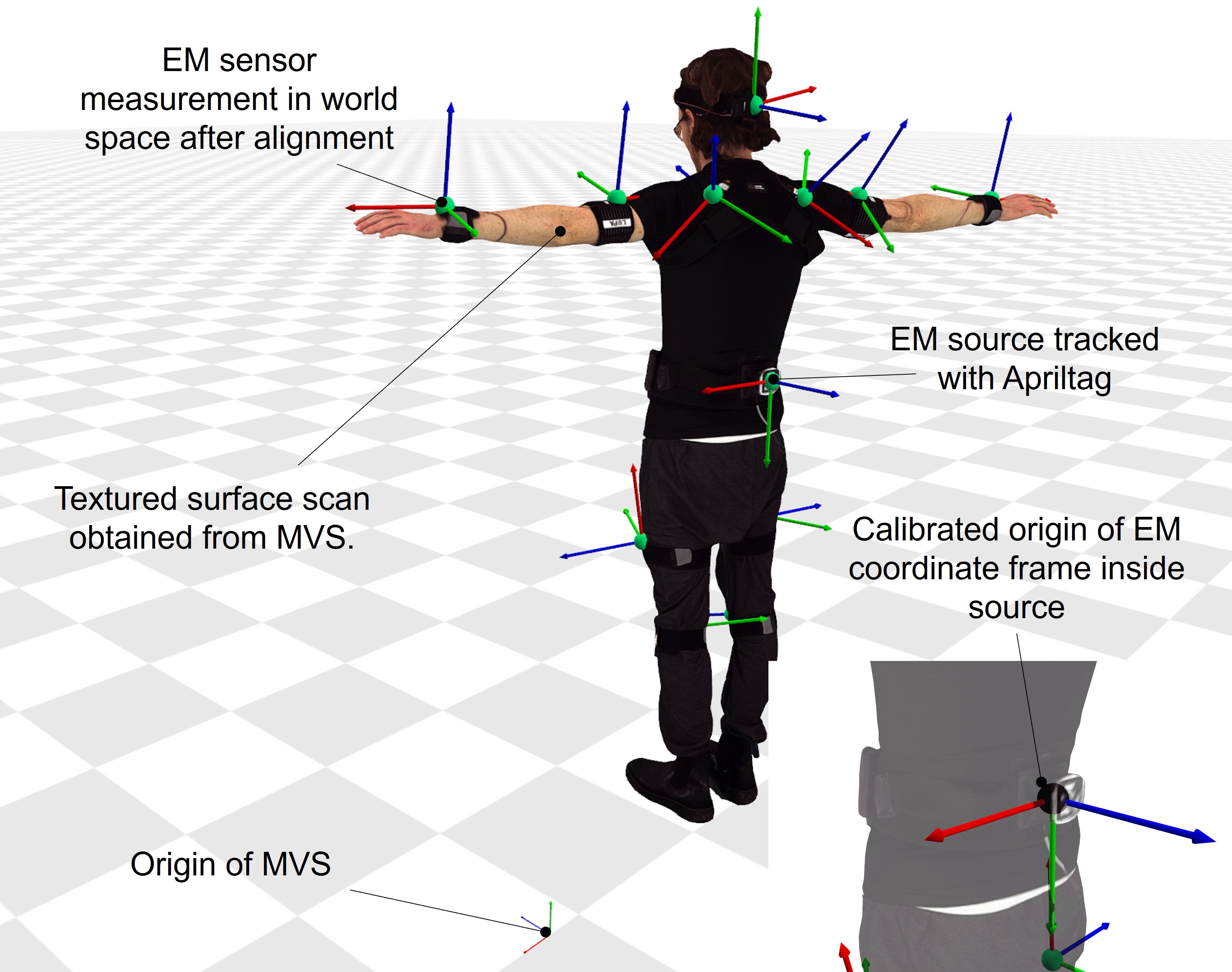}
    \caption{Visualization of coordinate frames and Apriltags involved for the body calibration procedure (see \secref{sec:supp_em_alignment}).}
    \label{fig:em_calibration}
\end{figure}

For the spatial alignment we track the EM source on the lower back with an Apriltag \cite{olson2011apriltag1,wang2016apriltag2,krogius2019apriltag3}. For a visualization please refer to \figref{fig:em_calibration}.
We track the EM source because the origin of the EM's coordinate system is the source. The Apriltag is a square with a side length of roughly 5 cm. With 53 RGB cameras with a resolution of $4\,088 \times 3\,000$ pixels we can triangulate the Apriltag's keypoints with millimeter accuracy. Only tracking the Apriltag is however not enough to align the coordinate frame of the MVS with the coordinate frame of the EM. This is because there is a constant rigid offset between the Apriltag and the center of the EM source where the origin of the EM coordinate frame is located. We thus need to determine this offset.

To do so we track an additional $S$ EM sensors with $S$ Apriltags and move the sensors around for roughly 15 seconds while recording both EM and image data. This allows us to formulate an optimization procedure that solves for the rigid constant offset between the Apriltag on the source and the unknown origin of the EM coordinate system.

For a calibration sequence of $T$ frames, let $\{ (\mathbf{p}_{s,t}^\text{E}, \mathbf{R}_{s,t}^\text{E}) \}_{t=1}^T$ be the position and orientation measurements for each sensor $1 \leqslant s \leqslant S$ in the EM-local coordinate system, \ie, relative to the source. Further, we assume time-synchronized 6-DoF Apriltag measurements $\{ (\mathbf{q}_{s,t}^\text{W}, \mathbf{U}_{s,t}^\text{W}) \}_{t=1}^T$ for each sensor $s$ in the world coordinate system, \ie, the MVS' coordinate frame. We denote the measurement of the Apriltag attached to the source with index $s = 0$. Here $\mathbf{q} \in \mathbb{R}^3$ and $\mathbf{U} \in SO(3)$.

Assuming an unknown rotational offset $\mathbf{R}_0^\text{o} \in SO(3)$ and an unknown translational offset $\mathbf{t}_0^\text{o} \in \mathbb{R}^3$ that describes the offset from the Apriltag on the source to the source center, we can compute the position of that origin in world coordinates as:

\begin{align}
    \mathbf{\dot{U}}_{0,t}^\text{W} &= \mathbf{U}_{0, t}^\text{W} \cdot \mathbf{R}_0^\text{o} \nonumber\\
    \mathbf{\mathbf{\dot{q}}_{0, t}^\text{W}} &= \mathbf{\dot{U}}_{0,t}^\text{W} \cdot \mathbf{t}_0^\text{o} + \mathbf{q}_{0, t}^\text{W} \label{eq:offsets}
\end{align}

We abbreviate \eqnref{eq:offsets} with a general function $\sigma(\mathbf{q}, \mathbf{U}, \mathbf{t}^\text{o}, \mathbf{R}^\text{o}$) that applies offsets $(\mathbf{t}^\text{o}, \mathbf{R}^\text{o})$ to positions and orientations $(\mathbf{q}, \mathbf{U})$. This is, we re-write \eqnref{eq:offsets}

\begin{equation}
    \mathbf{\mathbf{\dot{q}}_{0, t}^\text{W}}, \mathbf{\dot{U}}_{0,t}^\text{W} = \sigma(\mathbf{q}_{0, t}^\text{W}, \mathbf{U}_{0, t}^\text{W}, \mathbf{t}_0^\text{o}, \mathbf{R}_0^\text{o})
\end{equation}

Having determined the origin of the EM source in world space, \ie, $(\mathbf{\dot{q}}_{0, t}^\text{W}, \mathbf{\dot{U}}_{0,t}^\text{W})$ we can now map all EM sensor measurements into the world:

\begin{align}
    \mathbf{R}_{s,t}^\text{W} &=  \mathbf{\dot{U}}_{0,t}^\text{W} \cdot \mathbf{R}_{s,t}^\text{E} \nonumber \\
    \mathbf{p}_{s,t}^\text{W} &= \mathbf{\dot{U}}_{0,t}^\text{W} \cdot \mathbf{p}_{s,t}^\text{E} + \mathbf{\dot{q}}_{0, t}^\text{W} \label{eq:to_world}
\end{align}

As there is another rigid offset between the sensors' measurements in world space and the Apriltags attached to each sensor, we thus model another set of rigid rotational and translational offsets for every sensor $(\mathbf{t}_s^\text{o}, \mathbf{R}_s^\text{o})$ and compute

\begin{equation}
    \mathbf{\hat{p}}_{s,t}^\text{W}, \mathbf{\hat{R}}_{s,t}^\text{W} = \sigma(\mathbf{p}_{s,t}^\text{W}, \mathbf{R}_{s,t}^\text{W}, \mathbf{t}_s^\text{o}, \mathbf{R}_s^\text{o})
\end{equation}

\noindent With this, we formulate the objective:

\begin{equation}
    \argmin_{\mathcal{O}_s} \sum_{t=1}^T \sum_{s=1}^S
    ||\mathbf{\hat{p}}_{s,t}^\text{W} - \mathbf{q}_{s,t}^\text{W} ||^2_2 + ||  \mathbf{\hat{R}}_{s,t}^\text{W} - \mathbf{U}_{s,t}^\text{W} ||_2^2
\end{equation}
\noindent where $\mathcal{O}_s = \{(\mathbf{t}_s^\text{o}, \mathbf{R}_s^\text{o})\}_{s=0}^S$.
To sufficiently constrain this optimization we use $S=5$ sensors and move them around randomly for 15 seconds, \ie, $T=450$. The output of this spatial alignment are the offsets of the source $\mathcal{O}_0$. Note that the Apriltag is rigidly glued to the EM source, \ie, this procedure must only be done once.

\subsection{Computing Skin-To-Sensor Offsets}
With the source offsets $\mathcal{O}_0$ obtained in \secref{sec:supp_em_alignment} we can now move EM sensor measurements into the MVS' coordinate frame using \eqnref{eq:to_world}. This and our SMPL registration pipeline described in \secref{sec:supp_smpl_registration} allows us to compute skin-to-sensor offsets $\mathbf{o}_s = (\mathbf{v}_s, \mathbf{Q}_s)$ which are required for \methodname's stage 1.

To do so, we first define anchor points parameterized as a position $\mathbf{\tilde{p}}_s$ and orientation $\mathbf{\tilde{R}}_s$ on the SMPL mesh. The position $\mathbf{\tilde{p}}_s$ is simply the position of a hand-picked vertex and the orientation $\mathbf{\tilde{R}}_s$ can be constructed using any adjacent vertex and the corresponding vertex normal. Note that manually picking those anchor points on the SMPL mesh must only be done once.

Next, we take the registered SMPL meshes of a short 3-second calibration sequence and apply unknown offsets $(\mathbf{v}_s, \mathbf{Q}_s)$ to the anchor points to obtain virtual sensor orientations $\mathbf{R}_{s,t}^v = \mathbf{\tilde{R}}_{s,t} \mathbf{Q}_s$ and virtual sensor positions $\mathbf{p}_{s,t}^v = \mathbf{\tilde{R}}_{s,t}\mathbf{v}_s + \mathbf{\tilde{p}}_{s,t}$. We then equate the virtual measurements to the real measurements (which have been rotated to world space with $\mathcal{O}_0$) and optimize for the sensor offsets:

\begin{equation}
    \argmin_{\mathbf{v}_s, \mathbf{Q}_s} \sum_{t=1}^T || \mathbf{p}_{s,t}^\text{W} - \mathbf{p}_{s,t}^v||_2^2 + ||\mathbf{R}_{s,t}^\text{W} - \mathbf{R}_{s,t}^v||_2^2
\end{equation}

The output of this optimization are subject-specific skin-to-sensor offsets $\{\mathbf{o}_s \}_{s=1}^S$ which we compute for every participant and every capture session.

\section{Stage 3 Smoothing}
\begin{figure}
    \includegraphics[width=\linewidth, trim={0 0 0 0},clip]{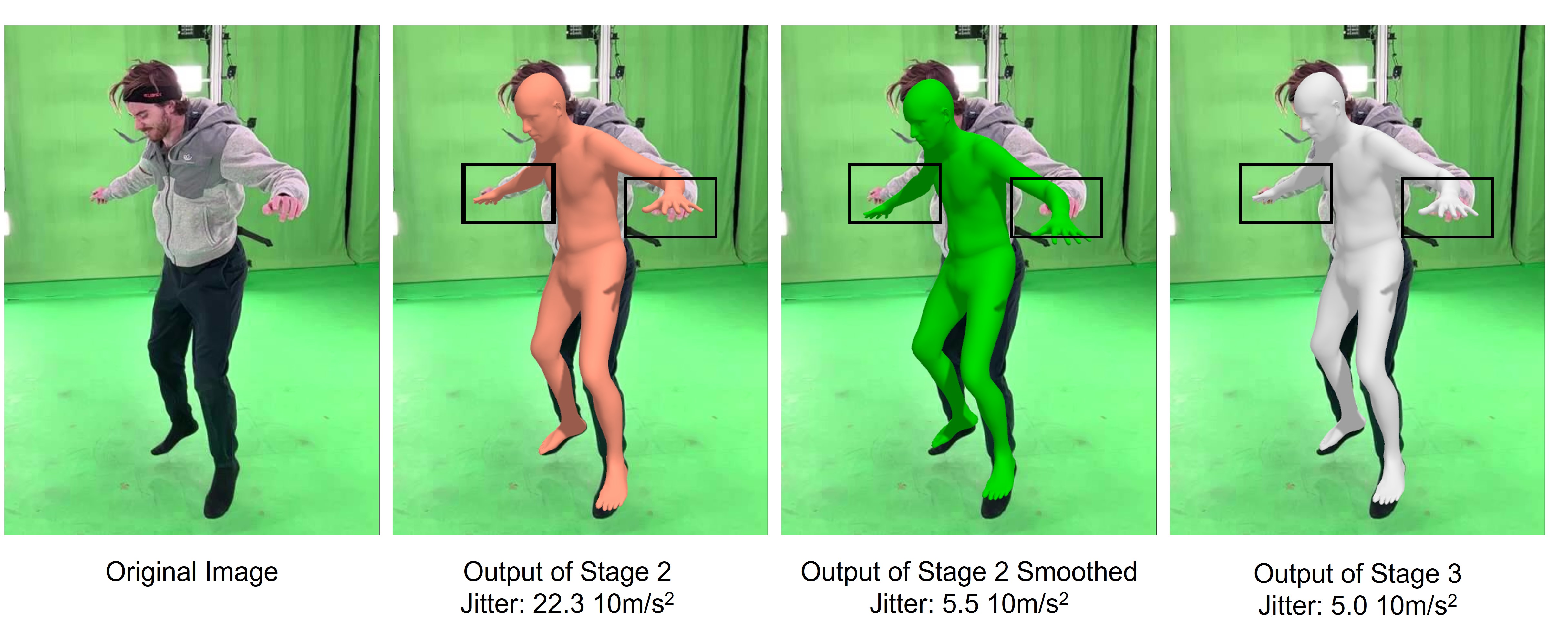}
    \caption{Effect of smoothing. The output of stage 2 (second to left) is jittery, but in this time instance well aligned with the image. Smoothing the output of stage 2 breaks this alignment (second to right). This is not the case for the outputs of stage 3 (right). Here, the pose is well aligned with the image observations while having a similar level of jitter as the na\"ively smoothed outputs of stage 2.}
    \label{fig:smoothing}
\end{figure}

As mentioned in the main paper in Sec.~6.1, we perform a light smoothing pass on the outputs obtained by minimizing $E_\text{S3}$. Because this only requires a light adjustment, it does not break pose-to-image alignment. Note that the same could not be said if we were to simply omit stage 3 and smooth the outputs of stage 2. To achieve the same reduction in jitter by smoothing stage 2, a more aggressive smoothing pass is required, which will lead to misalignments, especially in fast motions. We show and discuss such a case in \figref{fig:smoothing}.

In the smoothing pass, we smooth the SMPL parameters $\vectr{\theta}_r, \vectr{\theta}_b, \mathbf{t}$ using a Savitzky-Golay filter with a window length of $7$ and a second order polynomial. For the translation $\mathbf{t}$ we can directly apply the filter. For the SMPL body and root orientations, we first convert them into quaternions and apply the filter to each coordinate of the quaternions separately. For this to work, it is important to ensure that the quaternions are continuous because the quaternion $q$ represents the some rotation as the quaternion $-q$. Hence, we first make sure that the sign of a quaternion does not flip within a given sequence before we apply the filter. After the filter is applied, we normalize the quaternions to ensure valid rotations and convert them back to the angle-axis format.

\section{Visual Comparison to 3DPW}
\begin{figure}
    \includegraphics[width=\linewidth, trim={0 0 0 0},clip]{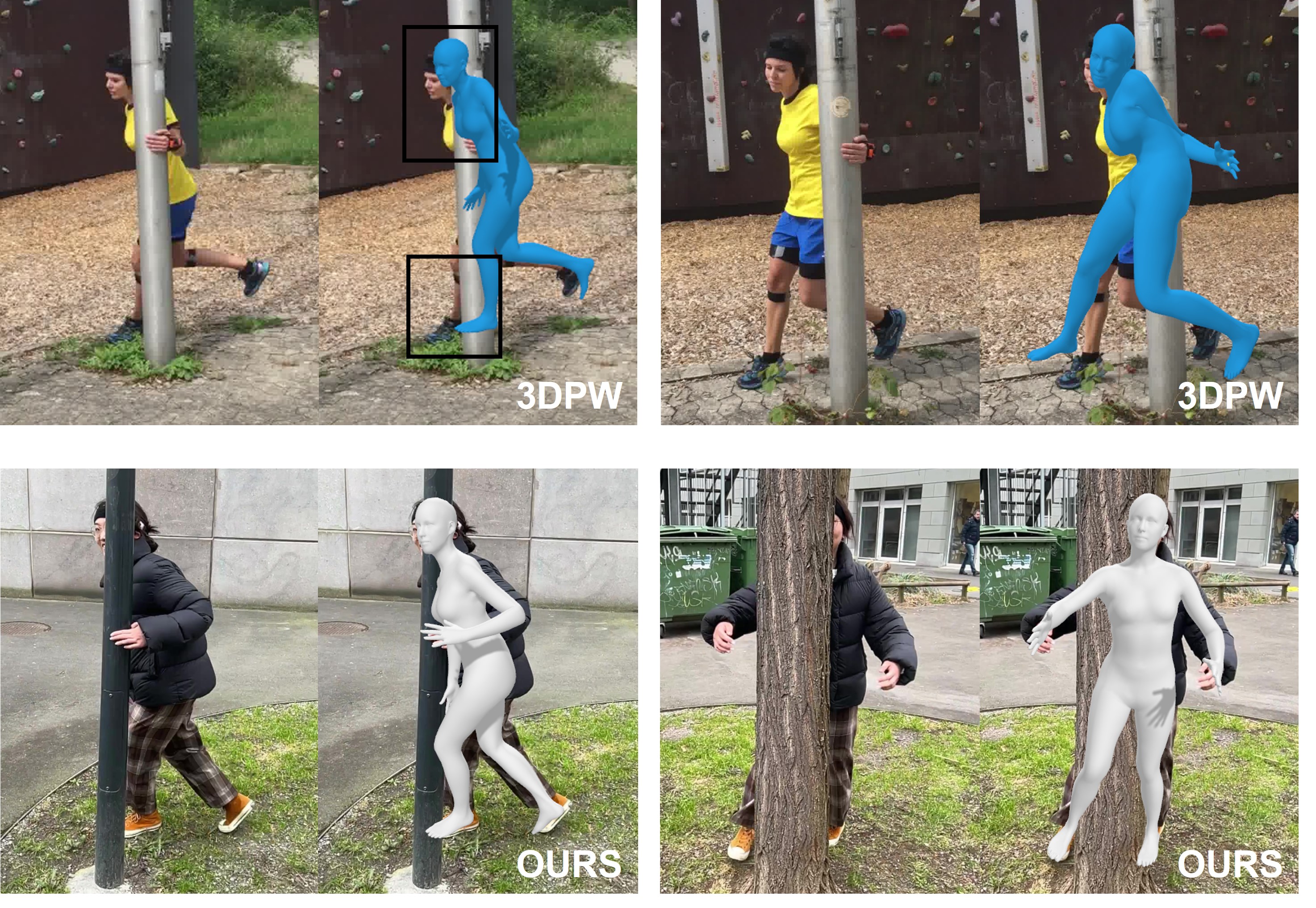}
    \caption{Visual comparison to 3DPW \cite{vonMarcard20183dpw} on a similar sequence. We observe sometimes large image-to-pose misalignments (top left) as well as unrealistic poses (top right) in 3DPW. In contrast, we provide better alignment and more accurate poses for similar (bottom left) or even higher levels of occlusion (bottom right).}
    \label{fig:comparison_3dpw}
\end{figure}

In the main paper we compare quantitately to 3DPW \cite{vonMarcard20183dpw} (see Sec.~6.1). Here we also show a few visual comparisons (see \figref{fig:comparison_3dpw}). To do so we recorded a similar motion sequence where the participant is walking around poles and is briefly occluded. In \figref{fig:comparison_3dpw} we observe higher fidelity SMPL fits in our results.

\section{Fine-tuning with \datasetname}
We fine-tune an existing human pose estimation method with \datasetname and investigate how this influences the performance on 3DPW (see \tabref{tab:emdb_finetuning}). For this example, we use ROMP \cite{Sun2021ROMP} and their publicly available code. The first row in \tabref{tab:emdb_finetuning} reports the result on 3DPW of their pre-trained model that has never seen 3DPW before. We observe that the joint errors on 3DPW decrease after fine-tuning this model with EMDB, which further higlights the usefulness of \datasetname.

\begin{table}
\centering
\begin{tabular}{lcc}
\toprule
Method & MPJPE & PA-MPJPE \\ \hline
ROMP (HRNet, w/o 3DPW) & 83.9 & 54.1 \\
ROMP + EMDB fine-tuning & \textbf{80.8} & \textbf{52.6}\\
\bottomrule
\end{tabular}
\caption{Effect of fine-tuning ROMP \cite{Sun2021ROMP} on \datasetname, evaluated on the test set of 3DPW, using ROMP's official, pre-trained model.}
\label{tab:emdb_finetuning}
\end{table}

\section{Evaluation of Global Trajectories}
\subsection{Camera Trajectory}
\label{sec:supp_camera_trajectory}
In Sec.~6.2 of the main paper we measure the accuracy of the iPhone's self-localized 6D poses. To do so, we attach an Apriltag rigidly to the iPhone and record both iPhone poses and Apriltag 6D poses on our MVS. This allows us to compare the iPhone's poses with the Apriltag tracking.
However, the former are in the iPhone's own coordinate system, while the latter are relative to the MVS' tracking space (because we triangulate the Apriltag with the known calibration of the MVS). In addition, there is a constant rigid offset between the iPhone's sensor origin and the Apriltag. We thus solve an optimization problem to align the two spaces, which is explained in the following.
Note that this problem is very similar to the optimization we run to align the EM space with the MVS as described in \secref{sec:supp_em_alignment}.

We are given triangulated Apriltag positions $\mathbf{q}_t^\text{W} \in \mathbb{R}^3$ and orientations $\mathbf{U}_t^\text{W} \in SO(3)$ in the MVS' coordinate system (here the world) and iPhone positions $\mathbf{p}^\text{i}_t \in \mathbb{R}^3$ and orientations $\mathbf{R}^\text{i}_t \in SO(3)$ in the iPhone's coordinate frame.

We first move the iPhone's 6D pose into the world with an unknown rigid transformation $\mathbf{T}^{\text{i} \rightarrow \text{W}} = \left[ \mathbf{R}^{\text{i} \rightarrow \text{W}} \mid \mathbf{t}^{\text{i} \rightarrow \text{W}} \right]$ to obtain

\begin{align}
    \mathbf{p}^\text{W}_t &= \mathbf{T}^{\text{i} \rightarrow \text{W}} \cdot \mathbf{p}^\text{i}_t \nonumber \\
    \mathbf{R}^\text{W}_t &= \mathbf{R}^{\text{i} \rightarrow \text{W}} \mathbf{R}^\text{i}_t
\end{align}

Next, we model an unknown translational and rotational offset $(\mathbf{t}^\text{o}, \mathbf{R}^\text{o})$ to account for the constant rigid offset between the Apriltag measurement and the iPhone's pose in the world space, \ie, $\mathbf{\hat{p}}_t^\text{W}, \mathbf{\hat{R}}_t^\text{W} = \sigma(\mathbf{p}_t^\text{W}, \mathbf{R}_t^\text{W}, \mathbf{t}^\text{o}, \mathbf{R}^\text{o})$ where $\sigma$ is the function defined in \secref{sec:supp_em_alignment}. We then compare the estimated Apriltag pose with the actual Apriltag measurement to minimize:

\begin{equation}
    \argmin_{\mathbf{T}^{\text{i} \rightarrow \text{W}}, \mathbf{t}^\text{o}, \mathbf{R}^\text{o}} \sum_{t=1}^T || \mathbf{\hat{p}}_t^\text{W} - \mathbf{q}_t^\text{W} ||^2_2 + ||  \mathbf{\hat{R}}_t^\text{W} - \mathbf{U}_t^\text{W} ||_2^2 
    \label{eq:iphone_min}
\end{equation}

\begin{figure}
    \includegraphics[width=\linewidth, trim={0 0 0 0},clip]{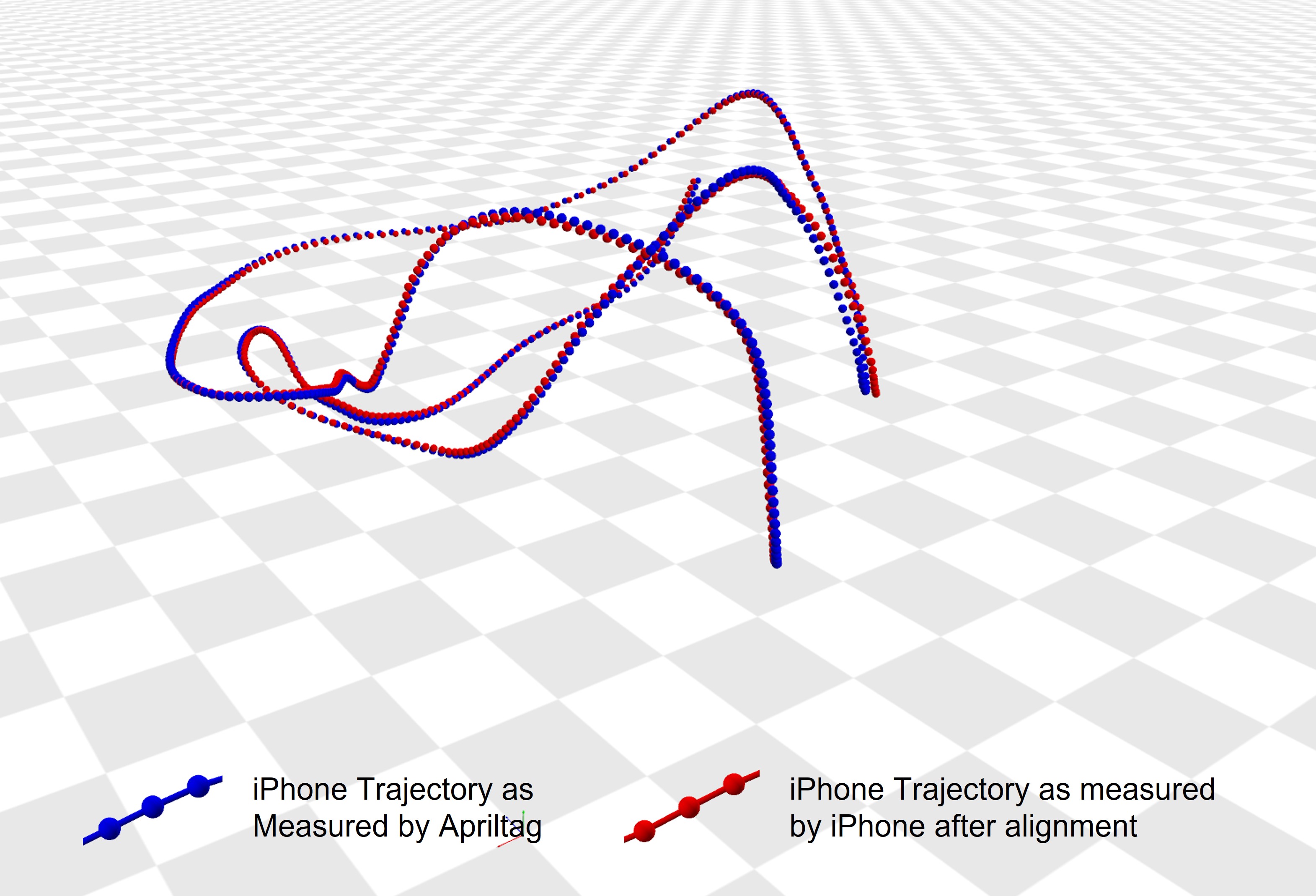}
    \caption{Visualization of optically tracked  (blue) and self-localized (red) iPhone trajectory.}
    \label{fig:iphone_trajectory}
\end{figure}

The objective value after this optimization is the alignment error we report in Sec.~6.2 (\textit{iPhone Pose Accuracy}) of the main paper. For a visualization of the aligned trajectories, please refer to \figref{fig:iphone_trajectory}.

\subsection{SMPL Root Trajectory}
We proceed similarly as described in \secref{sec:supp_camera_trajectory} to compute the error of the SMPL root trajectory estimated by \methodname to ground-truth SMPL root trajectories obtained with the MVS.
Note that we cannot simply re-use the transformation $\mathbf{T}^{\text{i} \rightarrow \text{W}}$ found in that section. This is because the iPhone's coordinate system changes with every new recording. In addition, evaluation takes with our MVS have fewer iPhone movements so as to not obstruct the MVS' cameras. This means that \eqnref{eq:iphone_min} tends to be underconstrained and thus the optimization does not always converge to meaningful solutions.

To address this, we add another term to \eqnref{eq:iphone_min} in which we move the SMPL root joint position predicted by EMP, $\mathbf{r}_t^\text{i}$, into the world frame and then compare it with the SMPL root joint given by our ground-truth registration, $\mathbf{r}_t^\text{W}$. This is, we compute $\mathbf{\hat{r}}_t^\text{W} = \mathbf{T}^{\text{i} \rightarrow \text{W}} \cdot \mathbf{r}_t^\text{i}$ and add the term $|| \mathbf{\hat{r}}_t^\text{W} - \mathbf{r}_t^\text{W} ||_2^2$ to \eqnref{eq:iphone_min}.
This effectively removes a global rigid misalignment between estimated and ground-truth SMPL root trajectories.
The remaining Euclidean distance between $\mathbf{\hat{r}}_t^\text{W}$ and $\mathbf{r}_t^\text{W}$, is the alignment error we report in the main paper in Sec.~6.2 (\textit{Global SMPL Trajectories}).

\section{\methodname Implementation Details}
We include implementation details of EMP's three stages here, as well as rough runtime estimates. We use PyTorch \cite{pytorch} for all computations.

In stage 1, we first only optimize for the SMPL root parameters $\vectr{\theta}_r, \mathbf{t}$ to get a rough alignment of the SMPL body to the sensor cloud. In a second pass we then optimize for all SMPL parameters, \ie, including $\vectr{\theta}_b$. For both passes, we use an L-BFGS \cite{DongNocedal1989LBFGS} optimizer with a learning rate of $1.0$ and strong Wolfe line search. We iterate the line search 20 times and take 5 steps with the L-BFGS optimizer. The remaining hyperparameters are chosen as $\lambda_\text{p} = 1.0, \lambda_\text{r} = 1.0, \lambda_\text{bp} = 1.0e^{-5}, \lambda_\text{rec} = 1.0$. This stage typically finishes in 1-2 minutes as we can use large batch sizes (the entire sequence fits into a single batch on a 24 GB GPU).

Stage 2 is a sequential optimization, where we optimize for each frame given the previous as initialization. In this stage we use the Adam optimizer \cite{kingma2014adam} with a learning rate of $0.01$. We optimize for $100$ iterations in each frame. The hyperparameters are set to $\lambda_\text{2D} = 0.01, \lambda_\text{rec} = 1.0, \lambda_\text{prior} = 1.0, \lambda_\text{pcl} = 10.0$. Each frame's optimization takes approx. $5$ seconds, so optimization of a typical sequence of $45$ seconds length finishes in roughly $2$ hours.

In stage 3 we fit a neural implicit human model. We also use the Adam optimizer \cite{kingma2014adam} with a learning rate of $5.0e^{-4}$. The learning rate decays to half after $200$ and $500$ epochs respectively. The hyperparameters are chosen as
$\lambda_\text{rgb} = 1.0, \lambda_\text{eik} = 0.1, \lambda_\text{reg} = 10.0$. The hyperparameters for the scene decomposition loss follow the same setting as in V2A \cite{Guo2023v2a}. Stage 3 is computationally the heaviest. We train the model for 48 hours on a single 24 GB GPU. We split very long sequences into several subparts and train each part in parallel on several GPUs in order to increase convergence speed.

\section{Socetial Impact}
Extracting human body shape and pose from imagery or other sensory data is an important building block in the endeavor to understand human behavior with computational methods. Having an accurate system available promises valuable applications such as immersive remote telepresence (thus saving CO2-intensive travel), automated rehabilitation (\eg, for the recovery of post-stroke patients), or computer-guided fitness and health coaches, to name just a few. All these applications directly benefit society, be it to provide more cost-effective treatments in medicine, smart tools to improve personal health, or ways to reduce our environmental footprint.

While this work does not directly improve human pose estimation methods, it fills an important gap: the availability of paired input sensor data and 3D human pose. This in turn will allow other researchers in the field to improve pose estimators by either using our dataset as training data or as a new evaluation benchmark. The evolution of Deep Learning in recent years has shown that the availability of data is a prime contributor to advancements in the field. Hence, we expect our dataset to lead to knowledge advancement in the field, which in turn will enable more sophisticated technical applications.

Human pose estimation methods, specifically so from images, might be abused for malicious surveillance or person identification via gait analysis or face recognition. Although \datasetname does not publish any identifiable information or directly propose improved pose estimators, future advancements in pose estimation directly imply that adverse uses of such technology automatically benefit, too. This presents an ethical and societal concern, which must be considered in future developments of such technology. Yet, given the promising technical applications of accurate human body pose estimators, we feel that the benefits of this research clearly outweighs the risks.

\end{document}